\documentclass{article} % For LaTeX2e
\usepackage{iclr2026_conference,times}
\usepackage{amsmath,amsfonts,bm}

\def\eqref#1{equation~\ref{#1}}
\def\1{\bm{1}}

\DeclareMathAlphabet{\mathsfit}{\encodingdefault}{\sfdefault}{m}{sl}
\SetMathAlphabet{\mathsfit}{bold}{\encodingdefault}{\sfdefault}{bx}{n}

\DeclareMathOperator*{\argmax}{arg\,max}
\DeclareMathOperator*{\argmin}{arg\,min}

\usepackage{hyperref}
\usepackage{url}
\hypersetup{
    colorlinks=true,
    citecolor=blue,
    linkcolor=blue,
    urlcolor=blue
}

\title{Exploring Image Generation via Mutually Exclusive Probability Spaces and Local Dependence Hypothesis}

\iclrfinalcopy

\author{Chenqiu Zhao\\
Department of Computer Science\\
University of Alberta\\
\texttt{\{zhao.chenqiu\}@ualberta.ca} \\
\And
Anup Basu\\
Department of Computer Science\\
University of Alberta\\
\texttt{\{basu\}@ualberta.ca} \\
}
\usepackage{dsfont}
\usepackage{graphicx}
\usepackage{wrapfig}
\usepackage{dsfont}
\usepackage{amssymb}
\usepackage{amsmath}
\usepackage{caption}
\usepackage{algorithm}
\usepackage{algorithmic}
\usepackage{tikz}
\usepackage{dsfont}

\usepackage[]{mdframed}

\usepackage{graphicx}
\usepackage{booktabs}
\usepackage{tabularx} % 导言区加
\usepackage{multirow}
\usepackage{mathtools}
\usepackage{makecell}
\usepackage{mathtools}
\usepackage{graphicx}

\usepackage{amssymb}
\usepackage{graphicx}
\usepackage{amsmath}
\usepackage{natbib}
\usepackage{comment}
\usepackage{amsthm}
\usepackage{amsmath}

\newcommand{\reffig}[1]{Fig. \ref{#1}}
\newcommand{\refsec}[1]{Sec. \ref{#1}}
\newcommand{\refthm}[1]{Theorem \ref{#1}}
\newcommand{\refdef}[1]{Definition \ref{#1}}
\newcommand{\refasm}[1]{Assumption \ref{#1}}

\newcommand{\reftab}[1]{Tab. \ref{#1}}
\newcommand{\refeqn}[1]{Eq. \ref{#1}}

\theoremstyle{definition}
\newtheorem{definition}{Definition}[section]

\newtheorem{theorem}[definition]{Theorem}
\newtheorem{assumption}[definition]{Assumption}

\newtheorem{remark}[definition]{Remark}



\usepackage{amsmath}
\let\argmin\relax
\let\argmax\relax

\DeclareMathOperator*{\argmin}{argmin}
\DeclareMathOperator*{\argmax}{argmax}

\usepackage{relsize}

\begin{document}

\maketitle

\begin{abstract}
A common assumption in probabilistic generative models for image generation is that learning the global data distribution suffices to generate novel images via sampling.
We investigate the limitation of this core assumption, namely that learning global distributions leads to memorization rather than generative behavior.
We propose two theoretical frameworks, the Mutually Exclusive Probability Space (MEPS) and the Local Dependence Hypothesis (LDH), for investigation.
	MEPS arises from the observation that deterministic mappings (e.g. neural networks) involving random variables tend to reduce overlap coefficients among involved random variables, thereby inducing exclusivity. 
	We further propose a lower bound in terms of the overlap coefficient, and introduce a Binary Latent Autoencoder (BL-AE) that encodes images into signed binary latent representations.
LDH formalizes dependence within a finite observation radius, which motivates our $\gamma$-Autoregressive Random Variable Model ($\gamma$-ARVM).
	$\gamma$-ARVM is an autoregressive model, with a variable observation range $\gamma$, that predicts a histogram for the next token. 
	Using $\gamma$-ARVM, we observe that as the observation range increases, autoregressive models progressively shift toward memorization. 
	In the limit of global dependence, the model behaves as a pure memorizer when operating on the binary latents produced by our BL-AE.
Comprehensive experiments and discussions support our investigation.
\end{abstract}

\section{Introduction}
\begin{wrapfigure}{r}{0.49\textwidth}
	\centering
	\vspace{-20pt}
	\includegraphics[width=0.99\linewidth]{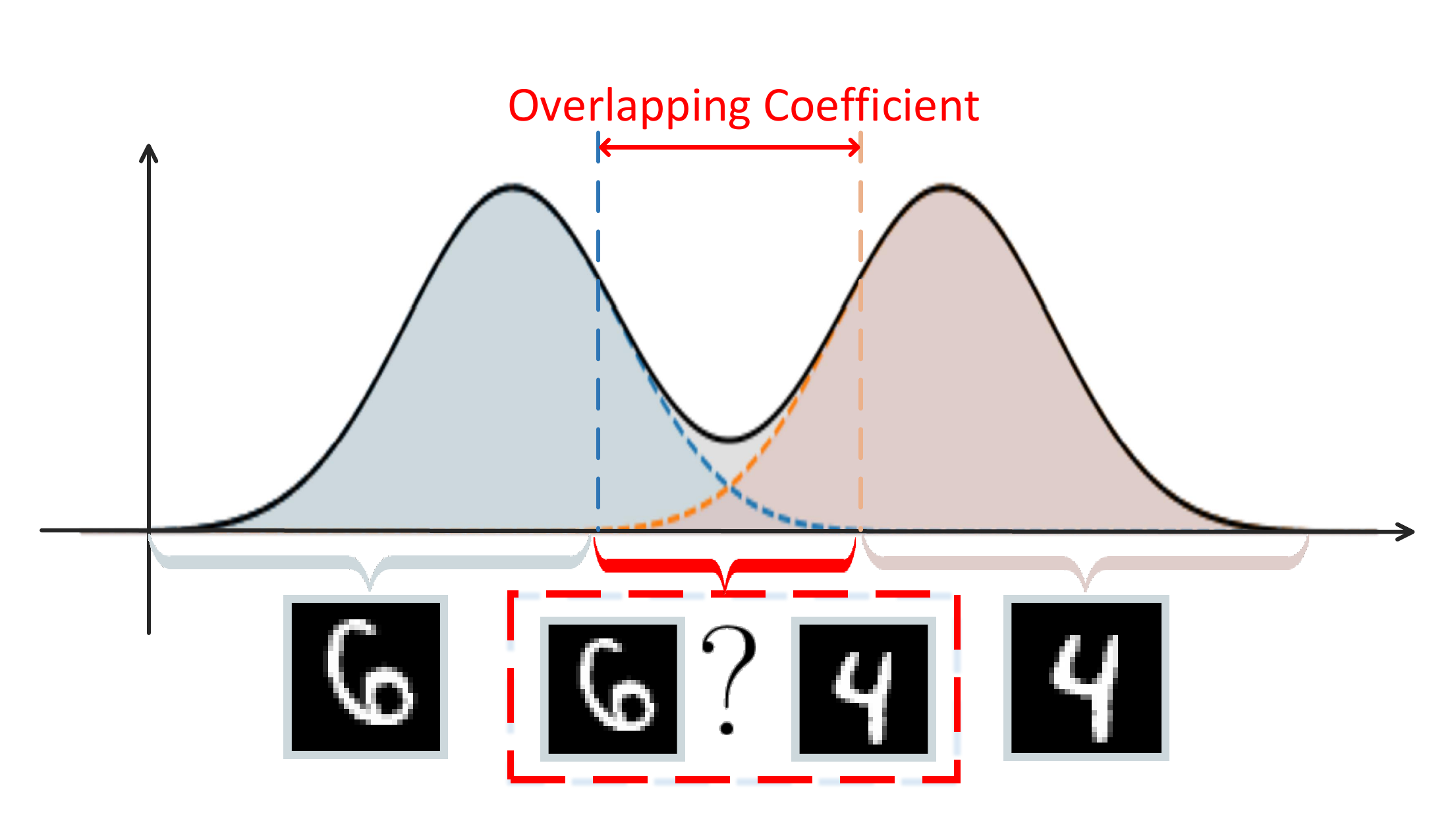}
	\caption{Selecting images for values in the overlap range is ambiguous.}
\label{fig_idea_v1}		
\end{wrapfigure}
Probabilistic generative models such as Variational Autoencoders (VAEs), Generative Adversarial Networks (GANs), diffusion models, and autoregressive models have achieved remarkable progress in image generation.
A core assumption is that these models learn an image distribution from which new images can be generated via sampling \citep{2022_TPAMI_9555209}.
However, we explore a potential limitation of this assumption, namely, that learning global distributions\footnote{By global distribution we mean the overall probability distribution that the generative model is trained to approximate across the entire dataset.} results in memorization rather than generative behavior.
Specifically, we focus on autoregressive models.
For this investigation, we introduce two theoretical frameworks.
The first, Mutually Exclusive Probability Space (MEPS),
arises from the observation that deterministic mappings involving random variables tend to reduce the overlap coefficients inherent to the system.
This reduction makes the probability spaces of the random variables effectively mutually exclusive.
The second is the Local Dependence Hypothesis (LDH),
which is motivated by an analysis of why autoregressive models tend to reproduce training samples.
While this phenomenon is often attributed to overfitting, we argue that it is related to the core assumption of learning global distributions.
The issue lies in differing philosophical views between the frequentist and Bayesian interpretations of whether probability distributions objectively exist.
This leads us to propose the Local Dependence Hypothesis (LDH), which posits that generative capacity in autoregressive models arises from modeling local dependence rather than global distributions.

In a trainable deterministic mapping from random variables to deterministic variables, for example, a network that takes noise as input for image reconstruction (like VAEs, GANs, or diffusion models),
the distributions of the random variables may overlap.
In such cases, observations from different optimization steps within the overlap region may be optimized toward inconsistent targets.
This is especially true when training for many epochs.
Consequently,
such inconsistent optimization targets raise the lower bound of the entire mapping system (\refthm{thm_lower_bound}), thereby degrading mapping fidelity (specifically, reconstruction quality).
As shown in \reffig{fig_idea_v1}, observations from overlapping ranges confuse the final optimization target.
When the random variables are also parameterized for optimization,
the learning dynamics tend to diminish such overlapping ranges,
and the means of these random variables are pushed apart (\refthm{thm_mutual}).
Exclusivity thus emerges.
This observation motivates the formulation of the Mutually Exclusive Probability Space (MEPS) (\refdef{def_MEPS}).
Leveraging this exclusivity,
we propose the Binary Latent Autoencoder (BL-AE), which encodes images into binary latent representations.
However, when feeding the learned binary latents into PixelCNN \citep{NIPS2016_b1301141}, a widely used autoregressive model,
the network often reproduces training samples.
This raises our concern that learning global distributions leads to memorization.
To investigate this possibility, we propose the Local Dependence Hypothesis (LDH),
which is formalized by assuming a bounded dependence radius for autoregressive models (\refasm{assume_LDH}).
Based on LDH, the $\gamma$-Autoregressive Random Variable Model ($\gamma$-ARVM) is proposed, which is an autoregressive model with a variable observation range $\gamma$.
In addition, given the subtle presence of MEPS in autoregressive models (\refsec{sec_auto_dis}), the proposed $\gamma$-ARVM outputs histograms describing the distribution of the next token rather than a label like PixelCNN.
The main contributions of this work are:
\begin{itemize}
	\item We propose the Mutually Exclusive Probability Space (\refdef{def_MEPS}) by observing exclusivity in an optimizable deterministic mapping system from random variables to deterministic targets.
		Based on this exclusivity, the Binary Latent Autoencoder (BL-AE) is introduced.
		In particular, by injecting noise into the outputs of activation functions with limited support width,
		the model learns signed binary latents, which are naturally used as tokens for autoregressive models.
		Moreover, MEPS can also be applied to revise the priors of generative models such as VAEs (\refsec{sec_vae_MEPS}) for improving fidelity.
	\item We propose the Local Dependence Hypothesis (LDH) (\refasm{assume_LDH}) to investigate a potential limitation in the core assumption of probabilistic generative models;
		namely, that learning global latent distributions may lead to memorization rather than generative behavior.
		In particular, the $\gamma$-Autoregressive Random Variable Model is proposed.
		Unlike previous autoregressive models that typically imply global dependence,
		the proposed $\gamma$-ARVM has a variable observation range $\gamma$.
		Using $\gamma$-ARVM, we observe that as the observation range increases, autoregressive models progressively shift toward memorization (\refsec{sec_exp_ldh}).
\end{itemize}
% Given the page limit, derivations and discussions that do not affect the main conclusions have been omitted. Please see the supplementary materials for full details.
\section{Related Work}
Probabilistic generative models have achieved remarkable progress across a range of applications.
A core assumption is that models learn a data distribution from which new content can be generated via sampling \citep{2022_TPAMI_9555209}.
For example,
Variational Autoencoders (VAEs) \citep{2014_VAE_Kingma2014} assume a Gaussian prior over latent variables and maximize the evidence lower bound (ELBO) to approximate the true posterior.
Generative Adversarial Networks (GANs) \citep{goodfellow2014generative} employ an adversarial objective, wherein a generator and a discriminator are trained in opposition.
Despite ongoing debate regarding whether GANs learn the true data distribution \citep{arora2018gans, chenminimax}, empirical results demonstrate the effectiveness of GANs in image generation \citep{Lee_Han_Kim_Choi_2025}.
Diffusion models \citep{ho2020denoising}, or score-based models \citep{songscore}, learn to generate data by reversing a diffusion process through score-function estimation.
Autoregressive models \citep{Chen_Pan_2025, Cheng_Yu_Tu_He_Chen_Guo_Zhu_Wang_Gao_Hu_2025} factorize the joint distribution into a product of conditionals and are usually combined with discrete latent quantization methods such as VQ-VAE \citep{NIPS2017_7a98af17}.
There are also other generative models such as energy-based models \citep{gao2021learning} and normalizing flows \citep{tabak2013family, papamakarios2019normalizing, 2022_PMLR_stimper22a, NEURIPS2022_b6341525}.
Most of these models share a fundamental assumption that learning a global distribution—whether over data or latent representations—is often traced back to the manifold hypothesis \citep{bengio2013representation}.
In this work, we propose the Mutually Exclusive Probability Spaces (MEPS) and the Local Dependence Hypothesis (LDH) to explore a potential limitation of this assumption.

Although our MEPS framework is newly proposed, the underlying principle can be observed in several previous works.
For example, the inconsistent optimization target is related to the optimization inconsistency in $\beta$-VAE \citep{higgins2017beta}, where the weights of the KL loss and the reconstruction loss are controlled by user-defined parameters.
Burgess et al.\ \citep{burgess2018understanding} explain this inconsistency through the information bottleneck, while Lucas et al.\ \citep{NEURIPS2019_7e3315fe} suggest that it leads to posterior collapse.
Recent work \citep{10_ijcai_2023_453} has also discussed the disentanglement of the reconstruction loss.
Moreover,
the inconsistency can also be observed in the ``prior hole'' problem \citep{aneja2021contrastive, xiao2020vaebm, nalisnick2018deep},
considering that a single Gaussian prior is insufficient for modeling complex data distributions \citep{vahdat2020nvae}.
In contrast,
Gaussian Mixture VAEs (GMVAEs) \citep{dilokthanakul2016deep, yang2019deep, guo2020variational} replace the standard prior with a mixture of Gaussians,
which reduces the overlap between the distributions of different latent variables, thereby alleviating the optimization inconsistency.
In this work, we mathematically demonstrate the exclusivity of these probability spaces and propose the Binary Latent Autoencoder (BL-AE).

The Local Dependence Hypothesis (LDH) can be viewed as an extension or improvement of the global dependence implicitly assumed in most autoregressive models \citep{NIPS2016_b1301141}.
Typically, autoregressive models imply global dependence, since they factorize the joint distribution into full-context conditionals.
However, there are also autoregressive models that incorporate local patterns \citep{local_fr_transform,NEURIPS2021_9996535e}, most of which were proposed primarily to reduce computational complexity.
For example, Cao et al.\ \citep{NEURIPS2021_9996535e} proposed a Local Autoregressive Transformer that restricts attention regions to accelerate inference.
In contrast, our work is, to the best of our knowledge, the first to systematically argue that learning the global distribution can lead to memorization.
Unlike prior work that devises attack methods to extract training samples from pre-trained large models such as Stable Diffusion \citep{ross2025a, NEURIPS2021_eae15aab, kowalczuk2025privacy,kasliwal2025localizing,yu2025icas}, our LDH serves as a theoretical framework to examine this foundational assumption in autoregressive models.

\section{Mutually Exclusive Probability Spaces}
\subsection{Theoretical Foundations}
\begin{definition}[Mutually Exclusive Probability Space (MEPS)]
Let $\tilde{\mathbf{Z}}= \{\tilde{\mathbf{z}}_i\}_{i=1}^N$ be a set of random variables with densities
$\{p_{\tilde{\mathbf{z}}_i}(\mathbf{z})\}_{i=1}^N$.
Let $\mathbf{X} = \{{\mathbf{x}}_i \}_{i=1}^{M}$ be a set of deterministic variables with $M \le N$.
For each pair $(i,j)$, the overlap coefficient between $\tilde{\mathbf{z}}_i$ and $\tilde{\mathbf{z}}_j$ is:
\begin{equation}
   \mathrm{OC}(\tilde{\mathbf{z}}_i,\tilde{\mathbf{z}}_j)
   = \int \min\!\big(p_{\tilde{\mathbf{z}}_i}(\mathbf{z}),\, p_{\tilde{\mathbf{z}}_j}(\mathbf{z})\big)\, d\mathbf{z}.
   \label{eqn_OC}
\end{equation}
Let $d_{\phi}:\tilde{\mathbf{Z}} \to \mathbf{X}$ be a deterministic mapping. We say that $(\tilde{\mathbf{Z}}, d_{\phi})$ forms a Mutually Exclusive Probability Space (MEPS) if:
\begin{equation}
   \max_{i\neq j}\; \mathrm{OC}(\tilde{\mathbf{z}}_i,\tilde{\mathbf{z}}_j) \;\le\; \varepsilon.
\end{equation}
When $\varepsilon=0$, we obtain a strict MEPS (all pairwise overlaps vanish up to measure zero).
When $\varepsilon>0$ is small, we obtain an approximate MEPS (pairwise overlaps are reduced to a negligible measure).
Note that strict MEPS is rare, unless otherwise stated, ``MEPS'' in this paper refers to the approximate case.
\label{def_MEPS}
\end{definition}
\begin{remark}
The MEPS definition can be understood as a characterization of overlap coefficients among random variables under a deterministic mapping, such as a neural network decoder. 
It also reflects a training objective: learning dynamics tend to reduce pairwise overlaps, 
thereby pushing the latent space closer to a strict MEPS. 
Thus, the definition serves both as a descriptive criterion and as a motivation for optimization. 
In addition, the densities $\{p_{\tilde{\mathbf{z}}_i}(\mathbf{z})\}_{i=1}^N$ must be parameterized by optimizable parameters.
Otherwise, overlaps remain fixed and cannot diminish. 
Note that MEPS still exist in this case, but only in a fixed form.
The deterministic mapping $d_{\phi}$ may be either trainable or fixed.
\end{remark}

\begin{theorem}[Reconstruction MSE Lower Bound]
\label{thm_lower_bound}
Let $\tilde{\mathbf{Z}}= \{\tilde{\mathbf{z}}_i\}_{i=1}^N$ be random variables with densities $\{p_{\tilde{\mathbf{z}}_i}\}_{i=1}^N$.
In particular, each random variable is obtained by injecting additive noise,
i.e., $\tilde{\mathbf{z}}_i=\mathbf{z}_i+\epsilon$, where for each $i$ the noise
$\epsilon$ is drawn independently from the same unimodal, symmetric distribution.
Let $\mathbf{X} = \{\mathbf{x}_i\}_{i=1}^N$ be deterministic targets.
Suppose $d_\phi:\tilde{\mathbf{Z}}\to {\mathbf{X}}$ is a deterministic mapping (e.g., a neural decoder). 
For each $i$, we reconstruct $\mathbf{x}_i$ as $d_\phi(\tilde{\mathbf{z}}_i)$ and evaluate the reconstruction error using the mean squared error (MSE).
Then the mean reconstruction loss satisfies:
\begin{equation}
\frac{1}{N}\sum_{i=1}^N \mathbb{E}_{\epsilon}\big[\|d_\phi(\tilde{\mathbf{z}}_i) - \mathbf{x}_i\|^2\big]
\;\ge\;
\frac{1}{4N^2}\sum_{i,j=1}^{N} \mathrm{OC}(\tilde{\mathbf{z}}_i,\tilde{\mathbf{z}}_j)\,\|\mathbf{x}_i-\mathbf{x}_j\|^2 .
\label{eqn_inequ_fin}
\end{equation}
The proof is provided in \refsec{sec_math_proof_lower_bound}.
\end{theorem}
\begin{remark}
This lower bound implies that the average reconstruction MSE cannot approach zero whenever the pairwise overlaps are nonzero.
Consequently, the reconstructed images cannot perfectly match the training targets under nonzero overlap.
In addition, although this lower bound is derived under the mean squared error,
a similar bound can be established for other convex loss functions.
This is because the constant term on the right-hand side arises from the convexity inequality and is independent of the overlap coefficient.
\end{remark}
\begin{theorem}[Mutual Exclusivity Theorem]
Let $p_{\tilde{\mathbf{z}}_i}$ and $p_{\tilde{\mathbf{z}}_j}$ be unimodal and symmetric 
densities centered at means $\mathbf{z}_i,\mathbf{z}_j \in \mathbb{R}^d$. 
Then minimizing the expectation of overlap coefficient satisfies:
\begin{equation}
      \argmin_{ {\mathbf{z}}_i, {\mathbf{z}}_j}\; 
   \mathbb{E}_{\epsilon} [\mathrm{OC}(\tilde{\mathbf{z}}_i,\tilde{\mathbf{z}}_j)]
   \Rightarrow  
   \argmin_{ {\mathbf{z}}_i, {\mathbf{z}}_j}\; 
   \mathbb{E}_{\epsilon} [\mathrm{OC}({\mathbf{z}}_i + \epsilon,{\mathbf{z}}_j + \epsilon)]
   \Rightarrow  
   \argmax_{ \mathbf{z}_i, \mathbf{z}_j }\; \frac{1}{N^2} \sum_{i,j}\|\mathbf{z}_i - \mathbf{z}_j\|^2 .
\end{equation}
	\label{thm_mutual}
	The proof is provided in \refsec{sec_math_proof_mutual}
\end{theorem}

\begin{remark}
Reducing pairwise overlaps under training dynamics forces the random variables to separate in expectation, thereby encouraging the formation of mutual exclusivity between random variables whose distributions exhibit overlap.
\end{remark}
\begin{wrapfigure}{r}{0.49\textwidth}
	\centering
	\scriptsize
	\resizebox{0.99\linewidth}{!}{ 
	\begin{tikzpicture}
		\tikzset{circle node/.style={
				circle,
				draw,
				minimum size=0.6cm,  
				text width=1.2cm,    
				align=center         
		}}
		\node[circle node] (VAE) at (90:2) {VAE};
		\node[circle node] (Diffusion) at (210:2) {Diffusion};
		\node[circle node] (GAN) at (330:2) {GAN};
		
		\draw[-latex] (Diffusion) to[bend right=20] node[midway, left, sloped, above] {Variable OC} (VAE);
		\draw[-latex] (VAE) to[bend right=20] node[midway, right, sloped, above] {Fixed OC} (Diffusion);
		
		\draw[-latex] (Diffusion) to[bend right=20] node[midway, below, sloped, below] {OC=1} (GAN);
		\draw[-latex] (GAN) to[bend right=20] node[midway, above, sloped, below] {Fixed OC} (Diffusion);
		
		\draw[-latex] (GAN) to[bend right=20] node[midway, right, sloped, above] {Variable OC} (VAE);
		\draw[-latex] (VAE) to[bend right=20] node[midway, left, sloped, above] {OC=1} (GAN);
	\end{tikzpicture}
	}
	\caption{VAE, GAN, and diffusion models are correlated by different assumptions introducing the overlap coefficient (OC) in the mutually exclusive probability space.}
	\label{fig_oe}		
\end{wrapfigure}
\textbf{MEPS in VAEs, diffusion models and GANs:}
MEPS exist in a wide range of probabilistic generative models, including VAEs, diffusion models, GANs, and even autoregressive models.
The presence of MEPS implies that reconstructed images cannot perfectly match training samples, which effectively prevents overfitting.
However, this also degrades reconstruction fidelity and thus leads to lower generation quality.
Since probabilistic generative models aim to approximate the data distribution, the trade-off between memorization and generalization becomes critical.
The overlap coefficient (OC), which characterizes MEPS, provides a natural measure of this trade-off.
As shown in \reffig{fig_oe}, diffusion models apply a fixed noise schedule, resulting in a fixed OC.
VAEs involve a competition between the KL term and the reconstruction loss, yielding a variable OC (typically less than 1).
In contrast, GANs sample directly from noise without explicit latent constraints, effectively corresponding to an OC of 1 (\refsec{sec_dis_gan_oc}).
As a result, diffusion models tend to be the easiest to train, while GANs are generally the most difficult.
Moreover, according to our \refthm{thm_lower_bound}, a lower overlap coefficient in the Mutually Exclusive Probability Space leads to lower reconstruction quality, which in turn typically results in better FID scores, due to the memorization .
Based on this observation, the fidelity of generated images (a better FID) can be improved by choosing priors that induce a lower overlap coefficient. 
For example, in VAEs, a Gaussian prior with a smaller variance $\sigma$ achieves better FID values (\refsec{sec_vae_MEPS}). 
While such behavior may reduce diversity, generative modeling is essentially a trade-off between fidelity and diversity.
Furthermore, by revising the prior assumptions of VAEs (\refsec{sec_dis_vae_mem}) , GANs (\refsec{sec_dis_gan_mem}), and diffusion models (\refsec{sec_dis_diffusion_mem}), one can deliberately drive these models toward memorization behavior, which highlights the significance of MEPS in studying the memorization properties of generative models.

\subsection{Binary Latent Autoencoder}
The Binary Latent Autoencoder is a practical application leveraging the exclusivity in MEPS.
We employ an activation function with a bounded output range, such as the hyperbolic tangent (tanh).
Then, noise from a symmetric bounded distribution is injected into the activation function's output, thereby extending the output into random variables,
with the motivation to form MEPS.
Due to exclusivity, the network implicitly pushes these latent variables toward distinct, non-overlapping regions at the limits of the tanh activation ($\pm 1$).
As training proceeds\footnote{Adding noise to the latent variables remains differentiable via the reparameterization trick, which allows gradients to pass through the stochastic sampling process.}, the activation outputs converge to binary values $\{-1,1\}$, resulting in an autoencoder with discrete signed binary latents.
Mathematically:
\begin{equation}
	\text{BL-AE}(\mathbf{x}_i;\theta, \phi) = \sum_{\mathbf{x}_i \in \mathbf{X}} \|d_{\phi}(\psi(e_{\theta}(\mathbf{x}_i))+ \sigma \cdot \epsilon)- \mathbf{x}_{i} \|^2,
\end{equation}
where $\psi(\cdot)$ is the activation function. $d_{\phi}, e_{\theta}$ denote the decoder and the encoder with $\phi$ and $\theta$ as their parameters.
$\mathbf{x}_i$ denotes an image from dataset $\mathbf{X}$.
$\epsilon$ is a noise following a distribution with unimodal and symmetric densities,
such the Gaussian distribution, or the generalized triangular distribution (GTD):
\begin{equation}
\epsilon \sim \mathrm{Tri}(\kappa)  = \left\{\begin{matrix}
	(1 - u^{\kappa}), \  \text{if} \ \  u > 0^{+} \\ (|u|^{\kappa} - 1), \ \text{if} \ \  u < 0^{-},
\end{matrix}\right.
\label{eqn_tri}
\end{equation}
where $u \sim \mathcal{U}(-1, 1)$ is the uniform distribution from -1 to 1.
The parameter $\kappa$ controls the sharpness of the distribution.
The proposed BL-AE works well to learn quantization tokens.
This is naturally suitable for autoregressive models.
We therefore, input the tokens from BL-AE into autoregression, 
and the memorization appears, which motivated us to propose the local dependence hypothesis for further investigation.

\section{Local Dependence Hypothesis}
\subsection{Theoretical Foundations}
\begin{assumption}[$\gamma$-Local Dependence Assumption ($\gamma$-LDA)]
Let $\{\tilde{\mathbf z}_i\}_{i=1}^N$ be random variables.
Fix a radius parameter $\gamma > 0$ under a given distance metric $d(\cdot,\cdot)$. 
We assume that the mutual information between variables is bounded by a tolerance $\varepsilon$:
\begin{equation}
   d(\tilde{\mathbf z}_i,\tilde{\mathbf z}_j) > \gamma  
   \;\;\Rightarrow\;\;
   I(\tilde{\mathbf z}_i;\tilde{\mathbf z}_j) \;\le\; \varepsilon .
\end{equation}
Thus, $\gamma$ defines a bounded dependence radius.
Beyond this radius, dependencies vanish up to $\varepsilon$-tolerance,
while within it, dependencies can be arbitrary.
When $\varepsilon=0$, we obtain a strict LDA, where exact independence holds outside radius $\gamma$.
When $\varepsilon>0$ is small, we obtain an approximate LDA, where long-range dependencies are reduced to negligible levels.
	\label{assume_LDH}
\end{assumption}
\begin{remark}
The $\gamma$-LDA hypothesis is conceptually related to the n-gram assumption in language modeling, as both impose locality by restricting the range of dependencies. 
	The n-gram assumption relies on a fixed-size window to truncate dependencies, primarily for natural language sequences. 
	In contrast, $\gamma$-LDA controls locality through mutual information with tolerance, making it applicable to more general settings such as images and other high-dimensional data. 
	Therefore, $\gamma$-LDA can be regarded as a generalization of the n-gram assumption.
Our $\gamma$-LDA is used to generalize autoregressive models. 
When the radius parameter $\gamma$ is greater than or equal to the sequence length, 
it reduces to standard autoregressive models such as PixelCNN \citep{NIPS2016_b1301141}. 
When the radius parameter $\gamma$ is smaller than the sequence length,
it effectively yields a local autoregressive model. 
Therefore, 
a variable observation range autoregressive model can be written as:
\begin{equation}
   p(\mathbf{Z}) = \prod_{i=1}^{N} p(\mathbf{z}_i \mid \mathbf{z}_{< i} )
   \;\;\Rightarrow\;\;
   p(\mathbf{Z}) = \prod_{i=1}^{N} p(\mathbf{z}_i \mid \mathbf{z}_{[i- \gamma,i)} ).
	\label{eqn_auto}
\end{equation}
\end{remark}

\subsection{$\gamma$-Autoregressive Random Variable Model}
Unlike previous common autoregressive models in image generation that imply global dependence,
the $\gamma$-Autoregressive Random Variable Model ($\gamma$-ARVM) is based on our $\gamma$-LDA, with variable observation ranges.
Moreover, the output of the proposed $\gamma$-ARVM also differs from regular masked architectures such as PixelCNN or Transformer,\footnote{Both rely on masking to enable parallel training in autoregressive models.} 
where both the input and the output are sequences of tokens. 
In our $\gamma$-ARVM,
the input is token sequences within an observation range $\gamma$. 
The output of the proposed $\gamma$-ARVM is a histogram that describes the distribution of the next token. 
Therefore, before training, the token distribution conditioned on the observation range $\gamma$ is captured in a statistical manner.
\begin{equation}
	q(\mathbf{Z}) \!=\! \prod_{i\!=\!1}^{N}q( \mathbf{z}_i \mid \mathbf{z}_{[i- \gamma,i)}) \!=\!\prod_{i\!=\!1}^{N} \mathbb{P}( \mathbf{z}_i\!=\!\mathbf{k} \mid \mathbf{z}_{[i- \gamma,i)} \!=\! \mathbf{g}) \!=\!\prod_{i\!=\!1}^{N} \frac{\sum_{n\!=\!1}^{N} \mathds{1}[\mathbf{z}_i^{(n)} \!=\! \mathbf{k}] \cdot \mathds{1}[\mathbf{z}_{[i- \gamma,i)}^{(n)} \!=\! \mathbf{g} ]}{\sum_{n\!=\!1}^{N} \mathds{1}[\mathbf{z}_{[i- \gamma,i)}^{(n)} \!=\! \mathbf{g} ]} ,
\end{equation}
where $\mathds{1}(\cdot)$ is the indicator function.
$\mathbf{k}$ and $\mathbf{g}$ are specific values of output token and input token sequences.
After capturing the training instances, the KL-divergence is used as the loss function:
\begin{equation}
	\mathcal{L}(p(\mathbf{Z}), q(\mathbf{Z}) ) =  \sum_{i = 1}^{N} \int q(\mathbf{z}| \mathbf{z}_{[i- \gamma,i)}) \log \frac{q(\mathbf{z} | \mathbf{z}_{[i- \gamma,i)})}{p(\mathbf{z}|\mathbf{z}_{[i- \gamma,i)})}  d\mathbf{z}.
\end{equation}
\subsubsubsection{\textbf{MEPS in Autoregressive Model:}}
\label{sec_auto_dis}
The main reason we do not follow the common sequence-to-sequence training paradigm is to handle the subtle MEPS in previous autoregressive models. 
From the mathematical definition of autoregression in the left part \refeqn{eqn_auto}, MEPS does not appear to exist. 
However, in practical implementations, MEPS inevitably emerges. The same input sequence may map to different next tokens, creating an inconsistent target that induces a non-vanishing lower bound for the cross-entropy loss. 
For example, consider the two binary sequences $(0,0,1,0)$ and $(0,0,0,0)$. 
From the first sequence, we obtain the mapping $(0,0)\rightarrow 1$, while from the second we obtain $(0,0)\rightarrow 0$.
This condition forces an overlap that cannot vanish, preventing the loss from approaching zero. 
Our proposed ARVM addresses this issue by predicting a histogram of the output label, e.g., 
$ (0,0)\rightarrow q(y \mid (0,0)) = [p_0,\, p_1] = [0.5,\, 0.5].$
In this way, the $\gamma$-ARVM can reduce the loss to extremely small values (e.g., $10^{-6}$), much smaller than those of sequence-to-sequence models such as PixelCNN. 
As a result, we are able to observe pure memorization conditions, which supports our claim that learning the global distribution tends to lead to memorization rather than genuine generative behavior.
While this phenomenon is frequently described as overfitting, one may argue that, in strict logical terms, the concept is somewhat redundant.
Because a near-zero loss naturally signifies optimization with respect to the chosen objective (\refsec{sec_dis_overfit}).

\section{Experiment}
\subsection{Mutually Exclusive Probability Space}
\subsubsection{Mutual Exclusivity Theorem}
\label{sec_exp_me}
\begin{figure}[th!]
	\centering
	\includegraphics[width=0.99\linewidth]{figure/fig_MEPS_v2.pdf}
	\vspace{-10pt}
	\caption{Demonstration of the mutual exclusivity theorem on the MNIST and CIFAR-10 datasets with Gaussian and triangular noise.}
	\label{fig_MEPS_exclu}
\end{figure}
We demonstrate the exclusivity in MEPS on MNIST and CIFAR-10 datasets with Gaussian and triangular noise settings,
with the motivation of supporting the generalization of MEPS.
In practice, the computation of overlap coefficient (OC) in high dimensions is extremely expensive.
Therefore, we adopt an indirect approach by scaling the variance parameter $\sigma$.
Increasing the value of $\sigma$ consistently increases the overlap for Gaussian and triangular distributions.
In particular, a straightforward autoencoder for image reconstruction with the architecture described in \reftab{tab_net_architect_ME} is utilized.
During training, we inject noise into the latent variables $\mathbf{z}_i$ to create random variables $\tilde{\mathbf{z}}_i$ following different distributions (Gaussian and Triangular).
Since mean squared error is used for reconstruction, the decoder becomes a deterministic mapping from latent random variables to deterministic targets (the ground-truth images).
MEPS thus emerges, and these latent random variables become mutually exclusive.
For demonstration, the MSE loss, the minimum distance between pairs of latents, and the average distance between the means of latent pairs are plotted in \reffig{fig_MEPS_exclu}.
Note that all curves are min–max normalized to $[0,1]$ for visualization, with normalization details provided in \reftab{table_details_norm}.
It is clear that the average distance between the means $\mathbf{z}_i$ and $\mathbf{z}_j$ of the latent variables $\tilde{\mathbf{z}}_i$ and $\tilde{\mathbf{z}}_j$ increases as the MSE decreases with the increasing number of training epochs.
This behavior is consistent with the Mutual Exclusivity Theorem (\refthm{thm_mutual}).

\subsubsection{Reconstruction MSE Lower Bound Theorem}
\begin{wrapfigure}{r}{0.49\textwidth}
	\centering
	\includegraphics[width=0.99\linewidth]{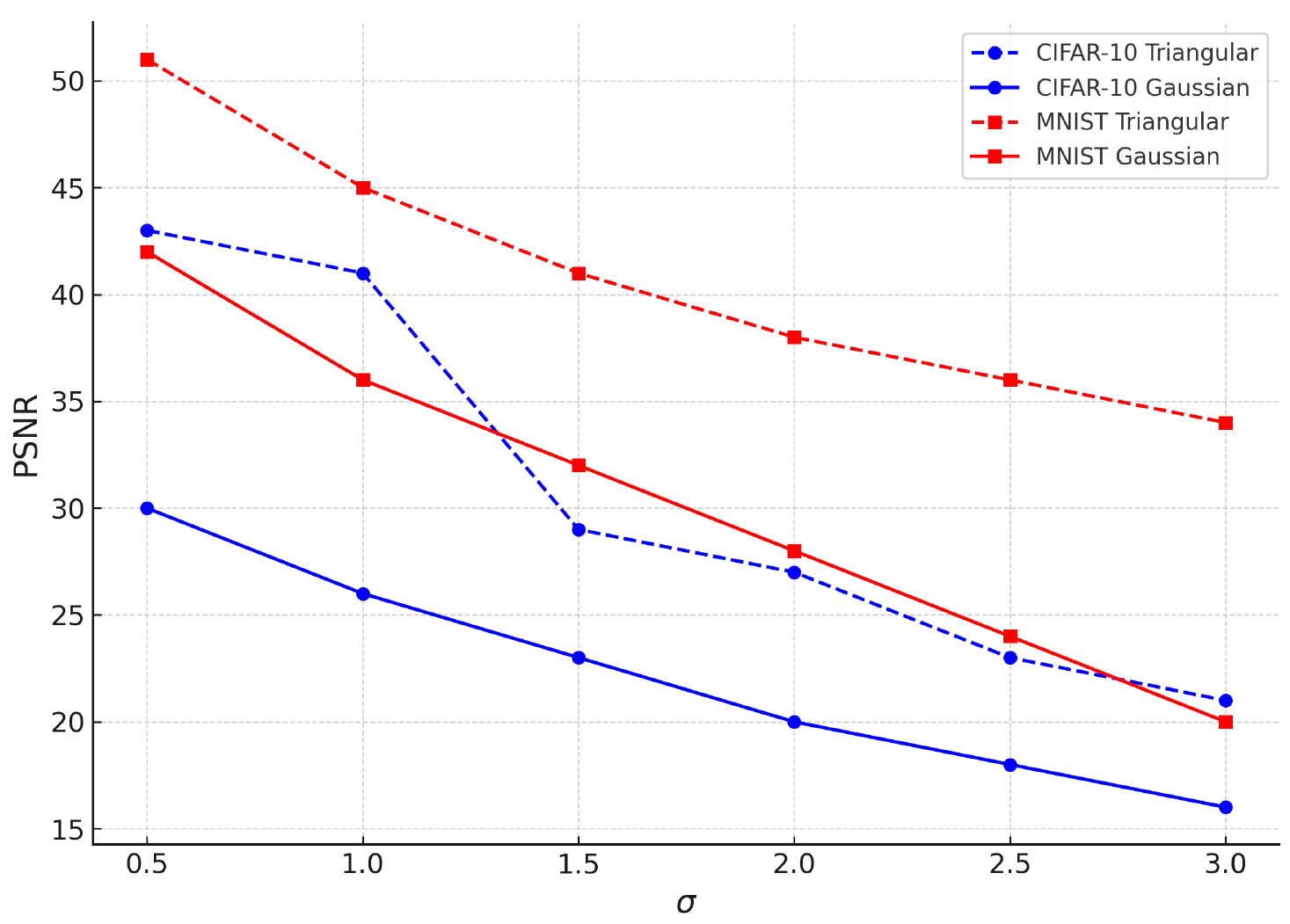}
	\vspace{-20pt}
	\caption{Ablation experiments of the lower bound with respect to the overlap coefficient, controlled by the noise standard deviation $\sigma$, on MNIST and CIFAR-10 with Gaussian and triangular noise.}
	\vspace{-30pt}
\label{fig_MEPS_sig_rec}		
\end{wrapfigure}
Moreover,
to further demonstrate the Reconstruction MSE lower bound,
we fixed the parameters of encoders in the previous section (\refsec{sec_exp_me}), 
and increased the intensity of noises for ablation experiments.
Specifically, 
we multiply the noise by a $\sigma$ value of $[0.5, 1.0, 1.5, 2.0, 2.5, 3.0]$,
and then retrain the decoder with a sufficient number of epochs.
When the value of $\sigma$ increases, the overlap coefficient increases as well.
Then based on \refthm{thm_lower_bound}, 
the lower bound of reconstruction error increases and leads to a decrease in reconstruction quality.
Therefore,
the plot of $\sigma$ and average reconstruction quality evaluated by Peak Signa-to-Noise Ratio (PSNR) is demonstrated in \reffig{fig_MEPS_sig_rec}.
It is clear to observe the monotonic trend that as $\sigma$ increases, the reconstruction quality decreases,
which is consistent with the Reconstruction MSE Lower Bound Theorem.
\subsubsection{Binary Latent Autoencoder}
\begin{figure}[th!]
	\centering
	\includegraphics[width=0.99\linewidth]{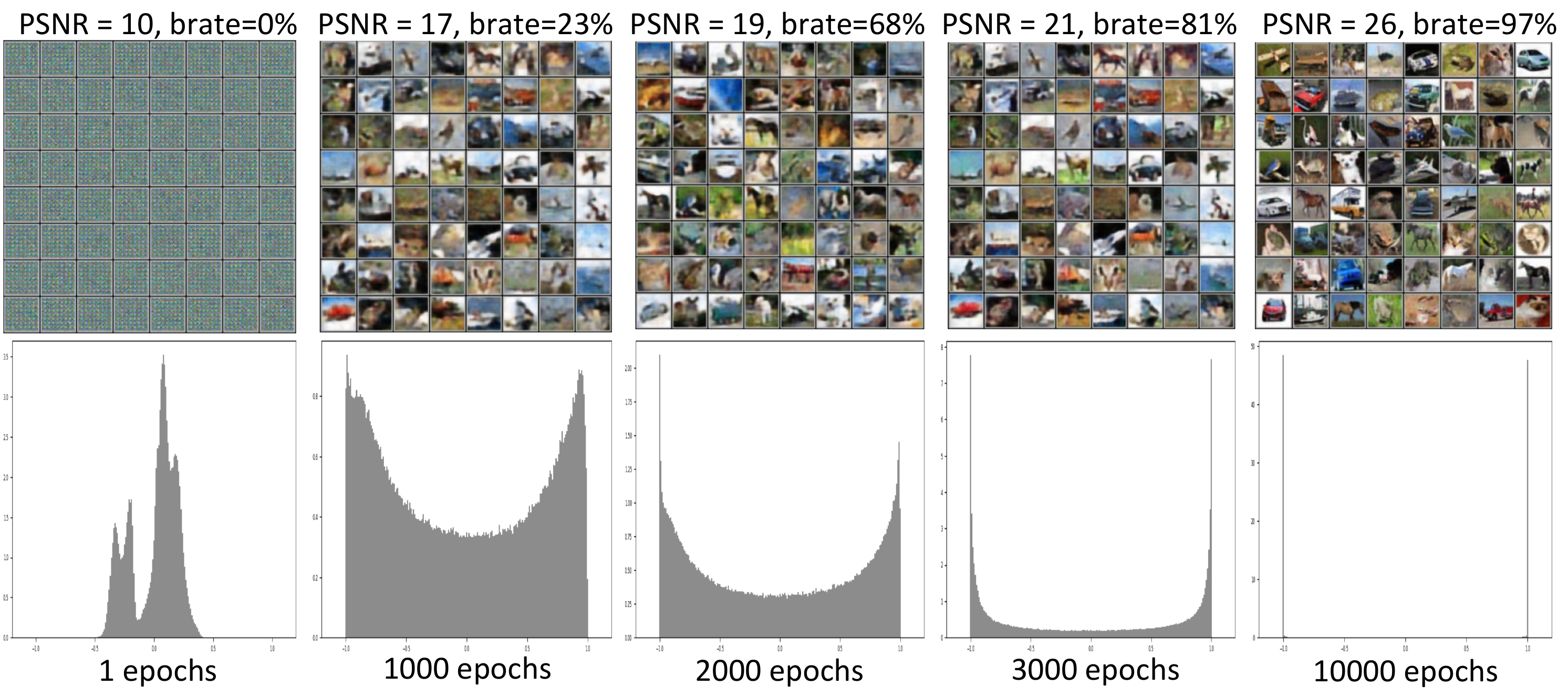}
	\caption{The demonstration of the binary rate of latent values (brate) and reconstruction quality under Peak Signal-to-Noise Ratio (PSNR) across training epochs. The histogram of latent values is illustrated on the second row.}
\label{fig_blae_hist}		
\end{figure}
\begin{figure}[htbp]
	\centering
	\includegraphics[width=0.99\linewidth]{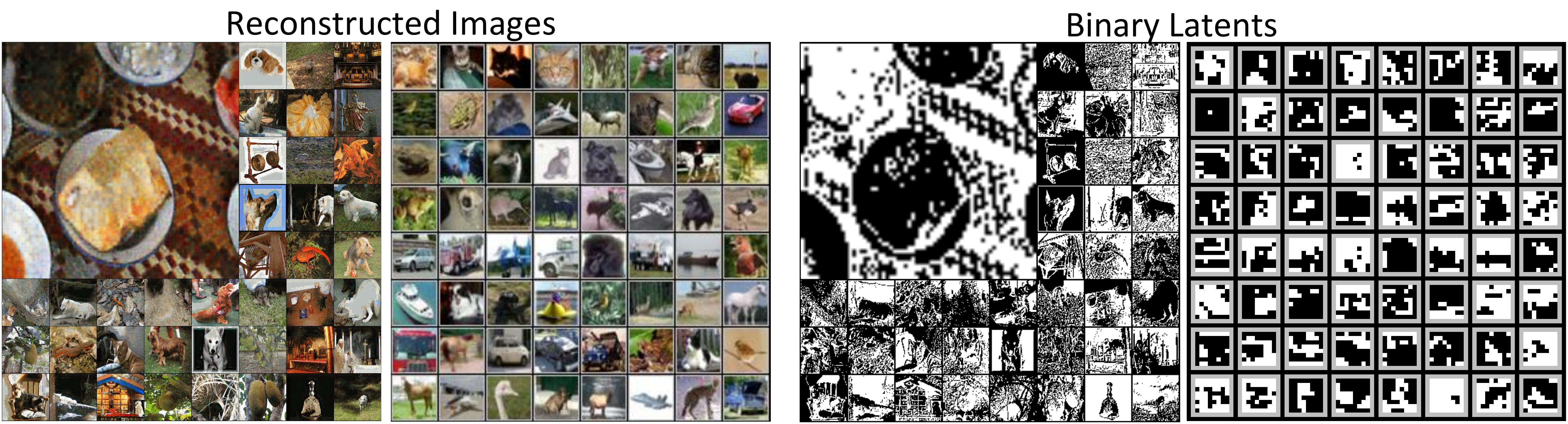}
	\vspace{-10pt}
	\caption{Qualitative evaluation of the proposed BL-AE on CIFAR-10 and 1k images subset of ImageNet.}
	\vspace{-20pt}
\label{fig_blae_vis}
\end{figure}
Our Binary Latent Autoencoder (BL-AE) has a few advantages compared to existing State-of-the-Art Autoencoders for latent representation extraction, such as VQ-VAE \citep{NIPS2017_7a98af17}, DC-AE \citep{DC_AE_abs_2410_10733}, and SD-VAE \citep{Rombach_2022_CVPR}.
First, BL-AE is able to capture discrete binary latent representations using a single reconstruction loss function (e.g., mean squared error).
In contrast, VQ-VAE relies on an additional K-means clustering to learn a codebook.
For demonstration, the values of latent variables converge to signed binary during training, as shown in \reffig{fig_blae_hist}.
This is because our BL-AE is based on the Mutually Exclusive Probability Space.
Second, the reconstruction quality of the proposed BL-AE correlates monotonically with the overlapping coefficient, which can be easily controlled by a parameter describing the intensity of noise, such as $\sigma$ (\reffig{fig_MEPS_sig_rec}).
Last but not least, as the latent values are binary, the latents provided by the proposed model require significantly less memory.
As shown in \reftab{tab_bin_cmp}, the total number of bits for DC-AE is $8 \times 8 \times 32 \times 32 = 65,536$ bits, whereas BL-AE only needs $16 \times 12 \times 1 = 192$ bits.
To further illustrate the visualization results, we encode subsets of the CIFAR and ImageNet datasets using latent sizes of $16 \times 16 \times 1$ and $64 \times 64 \times 1$, respectively.
The qualitative results of our binary latent representation are shown in \reffig{fig_blae_vis}.
\begin{table}[ht!]
\caption{Comparison between the proposed Binary Latent Autoencoder with state-of-the-art Autoencoders including DC-AE \citep{DC_AE_abs_2410_10733} and SD-VAE \citep{Rombach_2022_CVPR} in CIFAR-10 dataset.}
\label{tab_bin_cmp}
\begin{tabularx}{\linewidth}{l|X|r|r}
\toprule
\textbf{Method}  & \textbf{Latent Shape} & \textbf{rFID} $\downarrow$ & \textbf{PSNR}$\uparrow$ \\
\midrule
DC-AE$_{1}$ & $8\times8 \times 32 \times \text{32 bits}$ & 1.08 & 26.41 \\
DC-AE$_{2}$ & $4\times4\times128 \times \text{64 bits}$  & 2.30 & 28.71 \\
SD-VAE$_{1}$ & $8\times8 \times 32 \times \text{32 bits}$ & 6.81 & 19.01 \\
SD-VAE$_{2}$ & $4\times4\times128 \times \text{64 bits}$ & 8.53 & 22.34 \\
\midrule
Our BL-AE & $4\times4\times12 \times \text{1 bits}$ & 0.006 & 38.12 \\
\bottomrule
\end{tabularx}
	\vspace{-16pt}
\end{table}
\subsection{Local Dependence Hypothesis}
\label{sec_exp_ldh}
\begin{figure}[th!]
	\centering
	\includegraphics[width=0.99\linewidth]{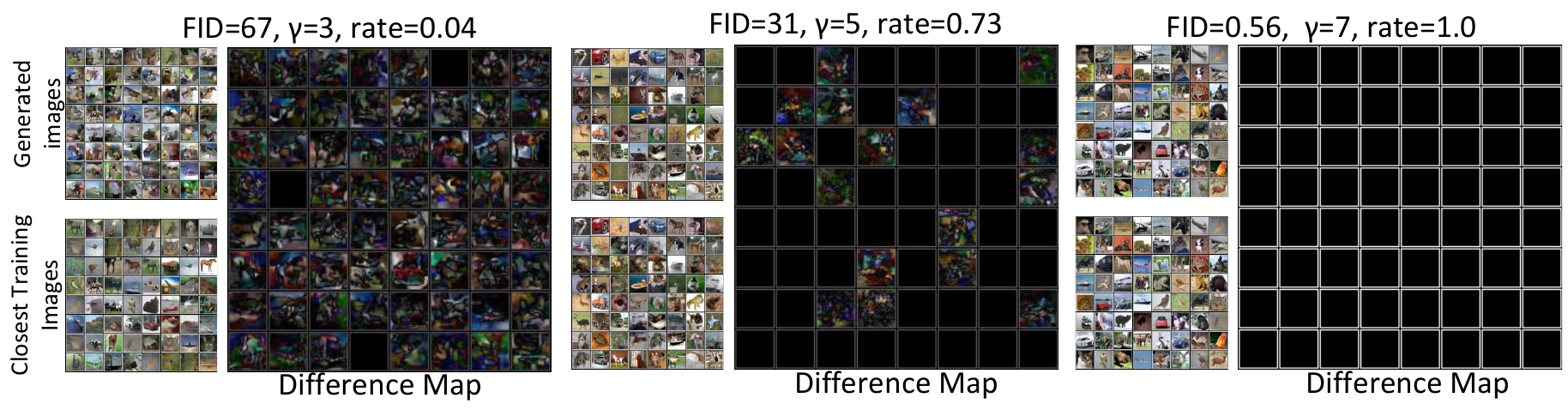}
	\vspace{-10pt}
	\caption{Images generated by our ARVM with observation range $\gamma$=7, 5, and 3.}
	\label{fig_global_mem}
	\vspace{-10pt}
\end{figure}
\begin{wrapfigure}{r}{0.49\textwidth}
	\centering
	\includegraphics[width=0.99\linewidth]{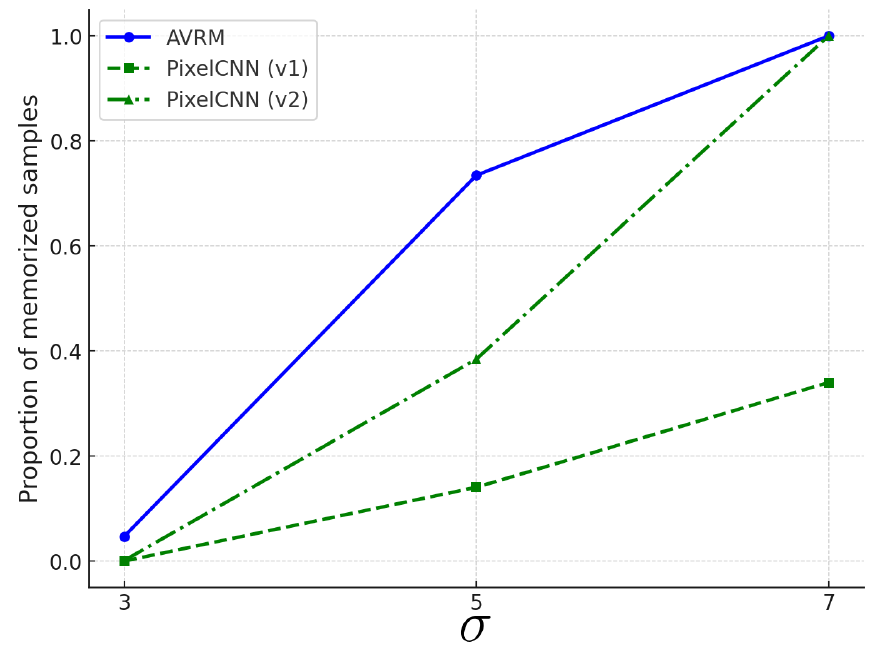}
	\vspace{-20pt}
	\caption{The demonstration of memorization rate with respect to the observation range $\gamma$ given values of 3, 5, 7.}
	\vspace{-30pt}
\label{fig_global_mem_plot}		
\end{wrapfigure}

The main focus of this paper is to investigate that learning global distribution leads to memorization.
Our LDH is proposed as the mathematical framework for our $\gamma$-ARVM.
The proposed $\gamma$-ARVM is able to learn distribution with a variable observation range $\gamma$.
We, therefore, have the theoretical and experimental tool for our investigation.
We first captured latents by our BL-AE with the architecture in \reftab{tab_net_bl_arvm},
with the latents of size $N \times 8 \times 4 \times 4$.
In particular, CIFAR-10 dataset is utilized.
Then by setting the observation ranges as 3, 5, and 7,
we observe that the ARVM becomes a pure observation model, which achieves very good FID values (\refsec{sec_cmp_ARVM}).
The results are shown in \reffig{fig_global_mem}.
For each observation range setting, we compute the memorization rate.
In particular, every generated image is computed PSNR with its closest training image.
When the PSNR value is greater than 30, we consider they are the same image, following common practice in image quality assessment.
Hence, we also utilized the PixelCNN \citep{NIPS2016_b1301141} to reproduce our experiments, with a popular implementation on GitHub.
The results are shown in \reffig{fig_global_mem_plot}.
In particular, when we first directly input the latents into PixelCNN, we did not observe a strong memorization,
which is the result of PixelCNN (v1).
We then double the network’s parameters and reduce the number of images to 1k, with the same architecture.
The PixelCNN (v2) also becomes a memorization model.
Since both the ARVM and PixelCNN achieve the same conclusion,
our concern that learning global distribution tends to lead to memorization rather than generative behavior is conducted.
We further evaluate memorization on high-resolution datasets, including 1k images from ImageNet and CelebA-HQ. 
Considering the computational cost, we directly increase the observation range to the global distribution. 
In this setting, memorization emerges prominently, with over 90\% memorization rate and competitive FID scores (\reftab{tab_comp_lsun_imagenet_celebahq}).
\subsection{On the Nature of Distributions}
The main reason for memorization in autoregressive models stems from a deeper philosophical question about the nature of probability distributions: what is a distribution? Is it an objective reality, or merely a subjective belief?
The frequentist perspective views probability distributions as objective realities, defined by the long-run frequencies with which events occur over time.
Thus, subjective prior assumptions should be strictly controlled, with minimal human intervention.
From this standpoint, memorization is not a flaw, but rather a faithful reflection of the empirical distribution observed from finite data, and arguably the best available approximation to the true distribution.
Autoregressive models embody this frequentist perspective.
However, in practical engineering, memorization is typically something to be avoided, and artificial prior assumptions are often introduced.
This aligns with the Bayesian view, which treats probability distributions as subjective beliefs.
Bayesian methods are therefore more flexible with prior assumptions, which is one of the main reasons why VAEs, GANs, and diffusion models employ a prior sampling distribution.
For example, Gaussian priors in VAEs and diffusion models.
Unfortunately, the reliance on prior assumptions introduces a high degree of subjectivity, and potentially even bias, into the evaluation and comparison of models.
Under such circumstances, a clear gap emerges between scientific objectivity and engineering subjectivity.
The proposed Local Dependence Hypothesis (LDH) can serve as a bridge for this gap.
Since locality is assumed, autoregressive models are still able to perform generation rather than memorization.

\section{Conclusion}
In this work, we proposed two theoretical frameworks: 1) Mutually Exclusive Probability Space (MEPS) and 2) the Local Dependence Hypothesis (LDH). 
These frameworks were designed to investigate a potential limitation in probabilistic generative modeling; namely, learning global distributions tends to result in memorization rather than true generation. 
In particular, 
we focus on autoregressive models.
MEPS motivated the development of the Binary Latent Autoencoder (BL-AE), which encodes images into binary latent representations. 
These representations serve as input to our Autoregressive Random Variable Model (ARVM), which can be configured to model either global distributions or local dependences.
When trained to model global distributions, ARVM becomes a memorization model.
In contrast, when local dependences are emphasized, 
ARVM exhibits generative behavior, producing novel images by recombining learned features. 
Comprehensive experiments and discussions were conducted to support our hypotheses.
\bibliography{iclr2026_conference}

\begin{thebibliography}{53}
\providecommand{\natexlab}[1]{#1}
\providecommand{\url}[1]{\texttt{#1}}
\expandafter\ifx\csname urlstyle\endcsname\relax
  \providecommand{\doi}[1]{doi: #1}\else
  \providecommand{\doi}{doi: \begingroup \urlstyle{rm}\Url}\fi

\bibitem[Aneja et~al.(2021)Aneja, Schwing, Kautz, and
  Vahdat]{aneja2021contrastive}
Jyoti Aneja, Alex Schwing, Jan Kautz, and Arash Vahdat.
\newblock A contrastive learning approach for training variational autoencoder
  priors.
\newblock \emph{Advances in neural information processing systems},
  34:\penalty0 480--493, 2021.

\bibitem[Arbel et~al.(2018)Arbel, Sutherland, Bi{\'n}kowski, and
  Gretton]{arbel2018gradient}
Michael Arbel, Danica~J Sutherland, Miko{\l}aj Bi{\'n}kowski, and Arthur
  Gretton.
\newblock On gradient regularizers for mmd gans.
\newblock \emph{Advances in neural information processing systems}, 31, 2018.

\bibitem[Arora et~al.(2018)Arora, Risteski, and Zhang]{arora2018gans}
Sanjeev Arora, Andrej Risteski, and Yi~Zhang.
\newblock Do gans learn the distribution? some theory and empirics.
\newblock In \emph{International conference on learning representations}, 2018.

\bibitem[Bengio et~al.(2013)Bengio, Courville, and
  Vincent]{bengio2013representation}
Yoshua Bengio, Aaron Courville, and Pascal Vincent.
\newblock Representation learning: A review and new perspectives.
\newblock \emph{IEEE transactions on pattern analysis and machine
  intelligence}, 35\penalty0 (8):\penalty0 1798--1828, 2013.

\bibitem[Bond-Taylor et~al.(2022)Bond-Taylor, Leach, Long, and
  Willcocks]{2022_TPAMI_9555209}
Sam Bond-Taylor, Adam Leach, Yang Long, and Chris~G. Willcocks.
\newblock Deep generative modelling: A comparative review of vaes, gans,
  normalizing flows, energy-based and autoregressive models.
\newblock \emph{IEEE Transactions on Pattern Analysis and Machine
  Intelligence}, 44\penalty0 (11), 2022.
\newblock \doi{10.1109/TPAMI.2021.3116668}.

\bibitem[Burgess et~al.(2018)Burgess, Higgins, Pal, Matthey, Watters,
  Desjardins, and Lerchner]{burgess2018understanding}
Christopher~P Burgess, Irina Higgins, Arka Pal, Loic Matthey, Nick Watters,
  Guillaume Desjardins, and Alexander Lerchner.
\newblock Understanding disentangling in beta -vae.
\newblock \emph{arXiv preprint arXiv:1804.03599}, 2018.

\bibitem[Cao et~al.(2021)Cao, Hong, Li, Wang, Xu, Fu, and
  Xue]{NEURIPS2021_9996535e}
Chenjie Cao, Yuxin Hong, Xiang Li, Chengrong Wang, Chengming Xu, Yanwei Fu, and
  Xiangyang Xue.
\newblock The image local autoregressive transformer.
\newblock In M.~Ranzato, A.~Beygelzimer, Y.~Dauphin, P.S. Liang, and J.~Wortman
  Vaughan (eds.), \emph{Advances in Neural Information Processing Systems},
  volume~34, pp.\  18433--18445. Curran Associates, Inc., 2021.

\bibitem[Carlini et~al.(2023)Carlini, Hayes, Nasr, Jagielski, Sehwag, Tramer,
  Balle, Ippolito, and Wallace]{carlini2023extracting}
Nicolas Carlini, Jamie Hayes, Milad Nasr, Matthew Jagielski, Vikash Sehwag,
  Florian Tramer, Borja Balle, Daphne Ippolito, and Eric Wallace.
\newblock Extracting training data from diffusion models.
\newblock In \emph{32nd USENIX Security Symposium (USENIX Security 23)}, pp.\
  5253--5270, 2023.

\bibitem[Chang et~al.(2022)Chang, Zhang, Jiang, Liu, and
  Freeman]{chang2022maskgit}
Huiwen Chang, Han Zhang, Lu~Jiang, Ce~Liu, and William~T Freeman.
\newblock Maskgit: Masked generative image transformer.
\newblock In \emph{Proceedings of the IEEE/CVF conference on computer vision
  and pattern recognition}, pp.\  11315--11325, 2022.

\bibitem[Chen et~al.(2024)Chen, Cai, Chen, Xie, Yang, Tang, Li, Lu, and
  Han]{DC_AE_abs_2410_10733}
Junyu Chen, Han Cai, Junsong Chen, Enze Xie, Shang Yang, Haotian Tang, Muyang
  Li, Yao Lu, and Song Han.
\newblock Deep compression autoencoder for efficient high-resolution diffusion
  models.
\newblock \emph{CoRR}, abs/2410.10733, 2024.
\newblock URL \url{https://doi.org/10.48550/arXiv.2410.10733}.

\bibitem[Chen et~al.(2022)Chen, Li, Li, and Meka]{chenminimax}
Sitan Chen, Jerry Li, Yuanzhi Li, and Raghu Meka.
\newblock Minimax optimality (probably) doesn't imply distribution learning for
  gans.
\newblock In \emph{International Conference on Learning Representations}, 2022.

\bibitem[Chen \& Pan(2025)Chen and Pan]{Chen_Pan_2025}
Zehao Chen and Rong Pan.
\newblock Svgbuilder: Component-based colored svg generation with text-guided
  autoregressive transformers.
\newblock \emph{Proceedings of the AAAI Conference on Artificial Intelligence},
  39\penalty0 (3):\penalty0 2358--2366, Apr. 2025.
\newblock \doi{10.1609/aaai.v39i3.32236}.

\bibitem[Cheng et~al.(2025)Cheng, Yu, Tu, He, Chen, Guo, Zhu, Wang, Gao, and
  Hu]{Cheng_Yu_Tu_He_Chen_Guo_Zhu_Wang_Gao_Hu_2025}
Kun Cheng, Lei Yu, Zhijun Tu, Xiao He, Liyu Chen, Yong Guo, Mingrui Zhu, Nannan
  Wang, Xinbo Gao, and Jie Hu.
\newblock Effective diffusion transformer architecture for image
  super-resolution.
\newblock \emph{Proceedings of the AAAI Conference on Artificial Intelligence},
  39\penalty0 (3):\penalty0 2455--2463, Apr. 2025.
\newblock \doi{10.1609/aaai.v39i3.32247}.

\bibitem[Cui et~al.(2023)Cui, Yu, Zhan, Liao, Lu, and Xing]{cui2023kd}
Kaiwen Cui, Yingchen Yu, Fangneng Zhan, Shengcai Liao, Shijian Lu, and Eric~P
  Xing.
\newblock Kd-dlgan: Data limited image generation via knowledge distillation.
\newblock In \emph{Proceedings of the IEEE/CVF Conference on Computer Vision
  and Pattern Recognition}, pp.\  3872--3882, 2023.

\bibitem[Dilokthanakul et~al.(2016)Dilokthanakul, Mediano, Garnelo, Lee,
  Salimbeni, Arulkumaran, and Shanahan]{dilokthanakul2016deep}
Nat Dilokthanakul, Pedro~AM Mediano, Marta Garnelo, Matthew~CH Lee, Hugh
  Salimbeni, Kai Arulkumaran, and Murray Shanahan.
\newblock Deep unsupervised clustering with gaussian mixture variational
  autoencoders.
\newblock \emph{arXiv preprint arXiv:1611.02648}, 2016.

\bibitem[et~al.(2021)]{esser2021taming}
Esser et~al.
\newblock Taming transformers for high-resolution image synthesis.
\newblock In \emph{Proceedings of the IEEE/CVF conference on computer vision
  and pattern recognition}, pp.\  12873--12883, 2021.

\bibitem[Gao et~al.(2021)Gao, Song, Poole, Wu, and Kingma]{gao2021learning}
R~Gao, Y~Song, B~Poole, YN~Wu, and DP~Kingma.
\newblock Learning energy-based models by diffusion recovery likelihood.
\newblock In \emph{International Conference on Learning Representations (ICLR
  2021)}, 2021.

\bibitem[Goodfellow et~al.(2014)Goodfellow, Pouget-Abadie, Mirza, Xu,
  Warde-Farley, Ozair, Courville, and Bengio]{goodfellow2014generative}
Ian Goodfellow, Jean Pouget-Abadie, Mehdi Mirza, Bing Xu, David Warde-Farley,
  Sherjil Ozair, Aaron Courville, and Yoshua Bengio.
\newblock Generative adversarial nets.
\newblock \emph{Advances in neural information processing systems}, 27, 2014.

\bibitem[Guo et~al.(2020)Guo, Zhou, Chen, Ying, Zhang, and
  Zhou]{guo2020variational}
Chunsheng Guo, Jialuo Zhou, Huahua Chen, Na~Ying, Jianwu Zhang, and Di~Zhou.
\newblock Variational autoencoder with optimizing gaussian mixture model
  priors.
\newblock \emph{IEEE Access}, 8:\penalty0 43992--44005, 2020.

\bibitem[Higgins et~al.(2017)Higgins, Matthey, Pal, Burgess, Glorot, Botvinick,
  Mohamed, and Lerchner]{higgins2017beta}
Irina Higgins, Loic Matthey, Arka Pal, Christopher Burgess, Xavier Glorot,
  Matthew Botvinick, Shakir Mohamed, and Alexander Lerchner.
\newblock beta-{VAE}: Learning basic visual concepts with a constrained
  variational framework.
\newblock In \emph{International Conference on Learning Representations}, 2017.

\bibitem[Ho et~al.(2020{\natexlab{a}})Ho, Jain, and
  Abbeel]{NEURIPS2020_4c5bcfec}
Jonathan Ho, Ajay Jain, and Pieter Abbeel.
\newblock Denoising diffusion probabilistic models.
\newblock In H.~Larochelle, M.~Ranzato, R.~Hadsell, M.F. Balcan, and H.~Lin
  (eds.), \emph{Advances in Neural Information Processing Systems}, volume~33,
  pp.\  6840--6851. Curran Associates, Inc., 2020{\natexlab{a}}.

\bibitem[Ho et~al.(2020{\natexlab{b}})Ho, Jain, and Abbeel]{ho2020denoising}
Jonathan Ho, Ajay Jain, and Pieter Abbeel.
\newblock Denoising diffusion probabilistic models.
\newblock \emph{Advances in neural information processing systems},
  33:\penalty0 6840--6851, 2020{\natexlab{b}}.

\bibitem[Karras et~al.(2018)Karras, Aila, Laine, and
  Lehtinen]{karras2018progressive}
Tero Karras, Timo Aila, Samuli Laine, and Jaakko Lehtinen.
\newblock Progressive growing of {GAN}s for improved quality, stability, and
  variation.
\newblock In \emph{International Conference on Learning Representations}, 2018.

\bibitem[Karras et~al.(2020)Karras, Aittala, Hellsten, Laine, Lehtinen, and
  Aila]{karras2020training}
Tero Karras, Miika Aittala, Janne Hellsten, Samuli Laine, Jaakko Lehtinen, and
  Timo Aila.
\newblock Training generative adversarial networks with limited data.
\newblock \emph{arXiv preprint arXiv:2006.06676v1}, 2020.

\bibitem[Kasliwal et~al.(2025)Kasliwal, Boenisch, and
  Dziedzic]{kasliwal2025localizing}
Aditya Kasliwal, Franziska Boenisch, and Adam Dziedzic.
\newblock Localizing and mitigating memorization in image autoregressive
  models.
\newblock \emph{arXiv preprint arXiv:2509.00488}, 2025.

\bibitem[Kingma \& Welling(2014)Kingma and Welling]{2014_VAE_Kingma2014}
Diederik~P. Kingma and Max Welling.
\newblock Auto-encoding variational bayes.
\newblock In \emph{International Conference on Learning Representations, 2014},
  2014.

\bibitem[Kowalczuk et~al.(2025)Kowalczuk, Dubi{\'n}ski, Boenisch, and
  Dziedzic]{kowalczuk2025privacy}
Antoni Kowalczuk, Jan Dubi{\'n}ski, Franziska Boenisch, and Adam Dziedzic.
\newblock Privacy attacks on image autoregressive models.
\newblock \emph{arXiv preprint arXiv:2502.02514}, 2025.

\bibitem[Lee et~al.(2025)Lee, Han, Kim, and Choi]{Lee_Han_Kim_Choi_2025}
Subeen Lee, Jiyeon Han, Soyeon Kim, and Jaesik Choi.
\newblock Diverse rare sample generation with pretrained gans.
\newblock \emph{Proceedings of the AAAI Conference on Artificial Intelligence},
  39\penalty0 (5):\penalty0 4553--4561, Apr. 2025.
\newblock \doi{10.1609/aaai.v39i5.32480}.

\bibitem[Li et~al.(2023)Li, Zhang, Li, Xu, Deng, and Li]{li2023neural}
Shengxi Li, Jialu Zhang, Yifei Li, Mai Xu, Xin Deng, and Li~Li.
\newblock Neural characteristic function learning for conditional image
  generation.
\newblock In \emph{Proceedings of the IEEE/CVF International Conference on
  Computer Vision}, pp.\  7204--7214, 2023.

\bibitem[Lucas et~al.(2019)Lucas, Tucker, Grosse, and
  Norouzi]{NEURIPS2019_7e3315fe}
James Lucas, George Tucker, Roger~B Grosse, and Mohammad Norouzi.
\newblock Don\textquotesingle t blame the elbo! a linear vae perspective on
  posterior collapse.
\newblock In H.~Wallach, H.~Larochelle, A.~Beygelzimer, F.~d\textquotesingle
  Alch\'{e}-Buc, E.~Fox, and R.~Garnett (eds.), \emph{Advances in Neural
  Information Processing Systems}, volume~32. Curran Associates, Inc., 2019.

\bibitem[Mao et~al.(2024)Mao, Wang, Xie, Li, and Wang]{local_fr_transform}
Xintian Mao, Jiansheng Wang, Xingran Xie, Qingli Li, and Yan Wang.
\newblock Loformer: Local frequency transformer for image deblurring.
\newblock In \emph{Proceedings of the 32nd ACM International Conference on
  Multimedia}, MM '24, pp.\  10382–10391, New York, NY, USA, 2024.
  Association for Computing Machinery.
\newblock ISBN 9798400706868.
\newblock \doi{10.1145/3664647.3680888}.

\bibitem[Michlo et~al.(2023)Michlo, Klein, and James]{10_ijcai_2023_453}
Nathan Michlo, Richard Klein, and Steven James.
\newblock Overlooked implications of the reconstruction loss for vae
  disentanglement.
\newblock In \emph{Proceedings of the Thirty-Second International Joint
  Conference on Artificial Intelligence}, IJCAI '23, 2023.
\newblock ISBN 978-1-956792-03-4.
\newblock \doi{10.24963/ijcai.2023/453}.

\bibitem[Nalisnick et~al.(2018)Nalisnick, Matsukawa, Teh, Gorur, and
  Lakshminarayanan]{nalisnick2018deep}
Eric Nalisnick, Akihiro Matsukawa, Yee~Whye Teh, Dilan Gorur, and Balaji
  Lakshminarayanan.
\newblock Do deep generative models know what they don't know?
\newblock In \emph{International Conference on Learning Representations}, 2018.

\bibitem[Papamakarios et~al.(2019)Papamakarios, Nalisnick, Rezende, Mohamed,
  and Lakshminarayanan]{papamakarios2019normalizing}
George Papamakarios, Eric Nalisnick, Danilo~Jimenez Rezende, Shakir Mohamed,
  and Balaji Lakshminarayanan.
\newblock Normalizing flows for probabilistic modeling and inference.
\newblock \emph{arXiv preprint arXiv:1912.02762}, 2019.

\bibitem[Parmar et~al.(2021)Parmar, Li, Lee, and Tu]{parmar2021dual}
Gaurav Parmar, Dacheng Li, Kwonjoon Lee, and Zhuowen Tu.
\newblock Dual contradistinctive generative autoencoder.
\newblock In \emph{Proceedings of the IEEE/CVF Conference on Computer Vision
  and Pattern Recognition}, pp.\  823--832, 2021.

\bibitem[Peebles \& Xie(2023)Peebles and Xie]{peebles2023scalable}
William Peebles and Saining Xie.
\newblock Scalable diffusion models with transformers.
\newblock In \emph{Proceedings of the IEEE/CVF International Conference on
  Computer Vision}, pp.\  4195--4205, 2023.

\bibitem[Rombach et~al.(2022)Rombach, Blattmann, Lorenz, Esser, and
  Ommer]{Rombach_2022_CVPR}
Robin Rombach, Andreas Blattmann, Dominik Lorenz, Patrick Esser, and Bj\"orn
  Ommer.
\newblock High-resolution image synthesis with latent diffusion models.
\newblock In \emph{Proceedings of the IEEE/CVF Conference on Computer Vision
  and Pattern Recognition (CVPR)}, pp.\  10684--10695, June 2022.

\bibitem[Ross et~al.(2025)Ross, Kamkari, Wu, Hosseinzadeh, Liu, Stein,
  Cresswell, and Loaiza-Ganem]{ross2025a}
Brendan~Leigh Ross, Hamidreza Kamkari, Tongzi Wu, Rasa Hosseinzadeh, Zhaoyan
  Liu, George Stein, Jesse~C. Cresswell, and Gabriel Loaiza-Ganem.
\newblock A geometric framework for understanding memorization in generative
  models.
\newblock In \emph{The Thirteenth International Conference on Learning
  Representations}, 2025.

\bibitem[Song \& Ermon(2019)Song and Ermon]{song2019generative}
Yang Song and Stefano Ermon.
\newblock Generative modeling by estimating gradients of the data distribution.
\newblock In \emph{Advances in Neural Information Processing Systems}, pp.\
  11895--11907, 2019.

\bibitem[Song et~al.(2020)Song, Sohl-Dickstein, Kingma, Kumar, Ermon, and
  Poole]{songscore}
Yang Song, Jascha Sohl-Dickstein, Diederik~P Kingma, Abhishek Kumar, Stefano
  Ermon, and Ben Poole.
\newblock Score-based generative modeling through stochastic differential
  equations.
\newblock In \emph{International Conference on Learning Representations}, 2020.

\bibitem[Stimper et~al.(2022)Stimper, Sch\"olkopf, and Miguel
  Hernandez-Lobato]{2022_PMLR_stimper22a}
Vincent Stimper, Bernhard Sch\"olkopf, and Jose Miguel Hernandez-Lobato.
\newblock Resampling base distributions of normalizing flows.
\newblock In \emph{Proceedings of The 25th International Conference on
  Artificial Intelligence and Statistics}, Proceedings of Machine Learning
  Research. PMLR, 2022.

\bibitem[Sun et~al.(2024)Sun, Jiang, Chen, Zhang, Peng, Luo, and
  Yuan]{sun2024autoregressive}
Peize Sun, Yi~Jiang, Shoufa Chen, Shilong Zhang, Bingyue Peng, Ping Luo, and
  Zehuan Yuan.
\newblock Autoregressive model beats diffusion: Llama for scalable image
  generation.
\newblock \emph{arXiv preprint arXiv:2406.06525}, 2024.

\bibitem[Tabak \& Turner(2013)Tabak and Turner]{tabak2013family}
Esteban~G Tabak and Cristina~V Turner.
\newblock A family of nonparametric density estimation algorithms.
\newblock \emph{Communications on Pure and Applied Mathematics}, 66\penalty0
  (2):\penalty0 145--164, 2013.

\bibitem[Vahdat \& Kautz(2020)Vahdat and Kautz]{vahdat2020nvae}
Arash Vahdat and Jan Kautz.
\newblock Nvae: A deep hierarchical variational autoencoder.
\newblock \emph{Advances in neural information processing systems},
  33:\penalty0 19667--19679, 2020.

\bibitem[van~den Burg \& Williams(2021)van~den Burg and
  Williams]{NEURIPS2021_eae15aab}
Gerrit van~den Burg and Chris Williams.
\newblock On memorization in probabilistic deep generative models.
\newblock In M.~Ranzato, A.~Beygelzimer, Y.~Dauphin, P.S. Liang, and J.~Wortman
  Vaughan (eds.), \emph{Advances in Neural Information Processing Systems},
  volume~34, pp.\  27916--27928. Curran Associates, Inc., 2021.

\bibitem[van~den Oord et~al.(2016)van~den Oord, Kalchbrenner, Espeholt,
  kavukcuoglu, Vinyals, and Graves]{NIPS2016_b1301141}
Aaron van~den Oord, Nal Kalchbrenner, Lasse Espeholt, koray kavukcuoglu, Oriol
  Vinyals, and Alex Graves.
\newblock Conditional image generation with pixelcnn decoders.
\newblock In D.~Lee, M.~Sugiyama, U.~Luxburg, I.~Guyon, and R.~Garnett (eds.),
  \emph{Advances in Neural Information Processing Systems}, volume~29. Curran
  Associates, Inc., 2016.

\bibitem[van den Oord~et al.(2017)]{NIPS2017_7a98af17}
van den Oord~et al.
\newblock Neural discrete representation learning.
\newblock In I.~Guyon, U.~Von Luxburg, S.~Bengio, H.~Wallach, R.~Fergus,
  S.~Vishwanathan, and R.~Garnett (eds.), \emph{Advances in Neural Information
  Processing Systems}, volume~30. Curran Associates, Inc., 2017.

\bibitem[Vuckovic()]{NEURIPS2022_b6341525}
James Vuckovic.
\newblock Nonlinear mcmc for bayesian machine learning.
\newblock In S.~Koyejo, S.~Mohamed, A.~Agarwal, D.~Belgrave, K.~Cho, and A.~Oh
  (eds.), \emph{Advances in Neural Information Processing Systems}, pp.\
  28400--28413. Curran Associates, Inc.

\bibitem[Wu et~al.(2019)Wu, Huang, Acharya, Li, Thoma, Paudel, and
  Gool]{wu2019sliced}
Jiqing Wu, Zhiwu Huang, Dinesh Acharya, Wen Li, Janine Thoma, Danda~Pani
  Paudel, and Luc~Van Gool.
\newblock Sliced wasserstein generative models.
\newblock In \emph{Proceedings of the IEEE/CVF Conference on Computer Vision
  and Pattern Recognition}, 2019.

\bibitem[Xiao et~al.(2020)Xiao, Kreis, Kautz, and Vahdat]{xiao2020vaebm}
Zhisheng Xiao, Karsten Kreis, Jan Kautz, and Arash Vahdat.
\newblock Vaebm: A symbiosis between variational autoencoders and energy-based
  models.
\newblock In \emph{International Conference on Learning Representations}, 2020.

\bibitem[Yang et~al.(2019)Yang, Cheung, Li, and Fang]{yang2019deep}
Linxiao Yang, Ngai-Man Cheung, Jiaying Li, and Jun Fang.
\newblock Deep clustering by gaussian mixture variational autoencoders with
  graph embedding.
\newblock In \emph{Proceedings of the IEEE/CVF international conference on
  computer vision}, 2019.

\bibitem[Yu et~al.(2025)Yu, Qiu, Yang, Fang, Zhuang, Hong, Chen, Wu, and
  Xia]{yu2025icas}
Hongyao Yu, Yixiang Qiu, Yiheng Yang, Hao Fang, Tianqu Zhuang, Jiaxin Hong, Bin
  Chen, Hao Wu, and Shu-Tao Xia.
\newblock Icas: Detecting training data from autoregressive image generative
  models.
\newblock \emph{arXiv preprint arXiv:2507.05068}, 2025.

\bibitem[Zhang et~al.(2025)Zhang, Zhao, Geng, Cohan, Luu, and
  Zhao]{zhang2025diffusion}
Siyue Zhang, Yilun Zhao, Liyuan Geng, Arman Cohan, Anh~Tuan Luu, and Chen Zhao.
\newblock Diffusion vs. autoregressive language models: A text embedding
  perspective.
\newblock \emph{arXiv preprint arXiv:2505.15045}, 2025.

\end{thebibliography}
\bibliographystyle{iclr2026_conference}

\newpage
\appendix
\section{Appendix}
\subsection{Discussion}

\subsubsection{Improving VAE with smaller standard deviation in Gaussian prior}
\label{sec_vae_MEPS}
\begin{figure}[th!]
\centering
\includegraphics[width=0.99\linewidth]{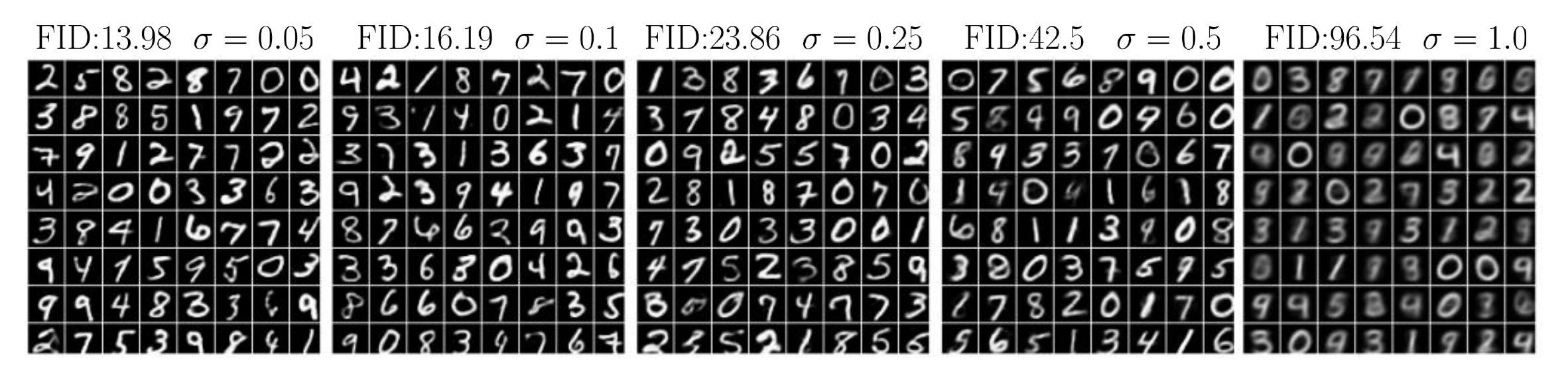}
\caption{The generation quality of Variational Autoencoder can be improved by using a smaller standard deviation $\sigma$ in the Gaussian prior.}
\label{fig_improve_VAE}
\end{figure}

Based on the proposed \refthm{thm_lower_bound},
the reconstruction quality is closely related to the lower bound with respect to the overlap coefficient.
An easy way to reduce the overlap coefficient is to use a smaller standard deviation $\sigma$,
since with decreasing $\sigma$, the latent spaces will have more room to tolerate the overlap coefficient.
To demonstrate this,
we utilized a standard VAE implementation from GitHub, and gradually reduced the $\sigma$ of noise.
Note that all other parts are kept invariant, including random seeds, optimization methods, training epochs, etc.
In particular, the MNIST dataset is utilized for evaluation.
As shown in \reffig{fig_improve_VAE},
the generation quality of the VAE increases with the decrease of $\sigma$.

\subsubsection{Memorization in Variational Autoencoder Models}
\label{sec_dis_vae_mem}
\begin{figure}[!t]
	\centering
	\includegraphics[width=0.99\linewidth]{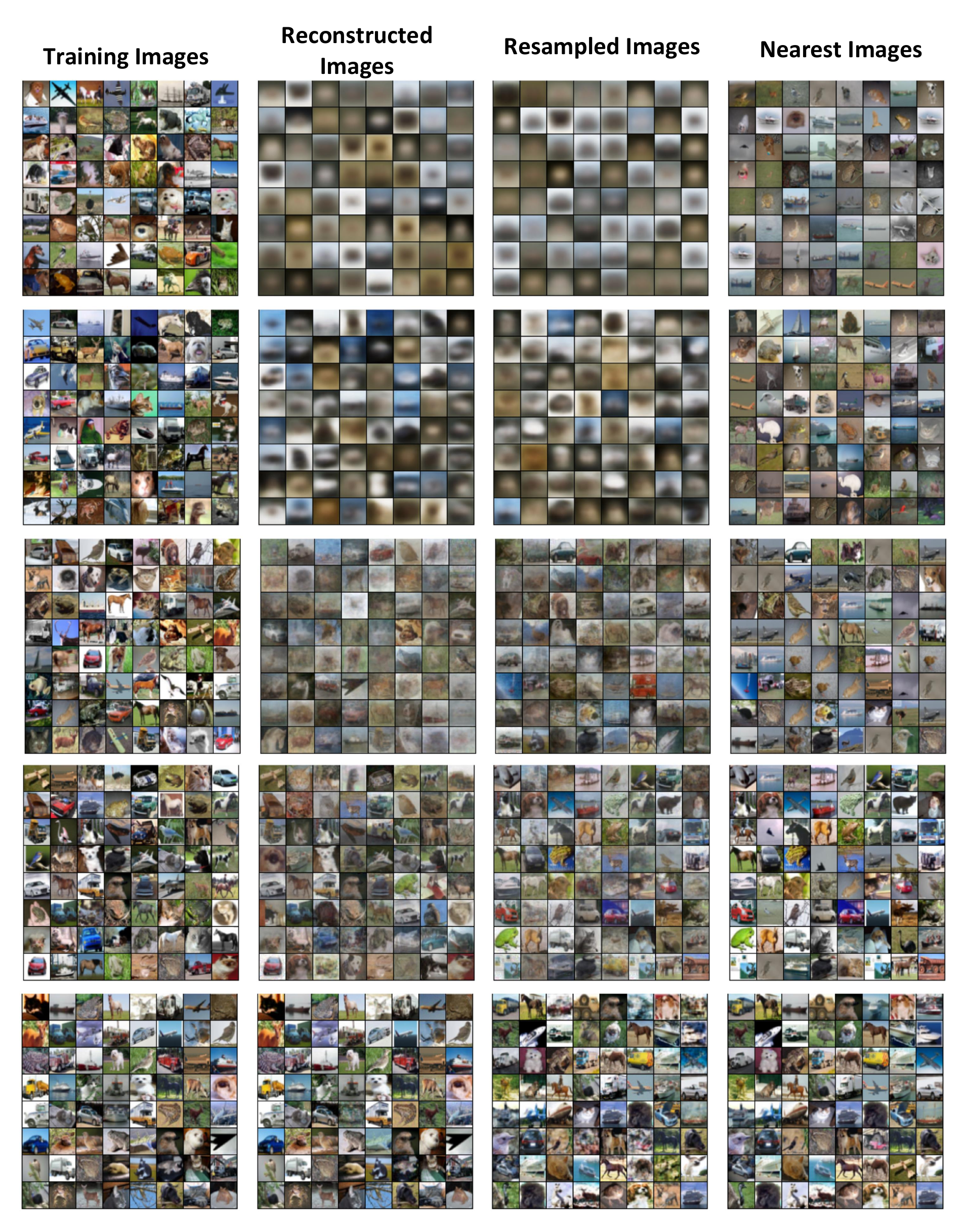}
	\DeclareGraphicsExtensions.
	\vspace{-0.6cm}
	\caption{By using a triangular distribution and limiting the values of latent samples in range $[-1, 1]$ it is easy to create continuous probabilistic fields in high dimensional space.
		All samples in this fields will generate images that are very similar to training images.}
	\label{fig_cifar}		
	%	\vspace{-0.4cm}
\end{figure}

The fundamental assumption of VAEs is to encode image distribution into a latent distribution, with the ELBO used for optimization.
The overlap coefficient in VAEs varies depending on the balance between the KL loss and the reconstruction loss.
To adjust the overlap coefficient, we replace Gaussian noise with triangular noise and constrain the latent variables using the tanh function, ensuring that values in the latent space remain within the range $[-1, 1]$.
Then by setting the $\sigma=1$, we create a condition that each dimension of latent space is able to include 2 different latent random variables without creating MEPS.
More specifically, two latent random variables centered at -1 and 1, with sigma as 1, so there are no overlap between the distributions of these two latents random variables.
With the increase in the number of dimensions, the total possible number of random variables without OC becomes $2^M$, where M is the number of dimensions.
As the number of dimensions increases, we obtain the results shown in \reffig{fig_cifar}.
In the low-dimensional case, since the total space for the overlap coefficient is insufficient, the reconstruction is not similar to the training images.
However, with an increasing number of dimensions, the reconstructed images gradually become closer to the training images.

\subsubsection{Memorization in Generative Adversarial Network}
\label{sec_dis_gan_mem}
\begin{figure}[!t]
	\centering
	\includegraphics[width=0.99\linewidth]{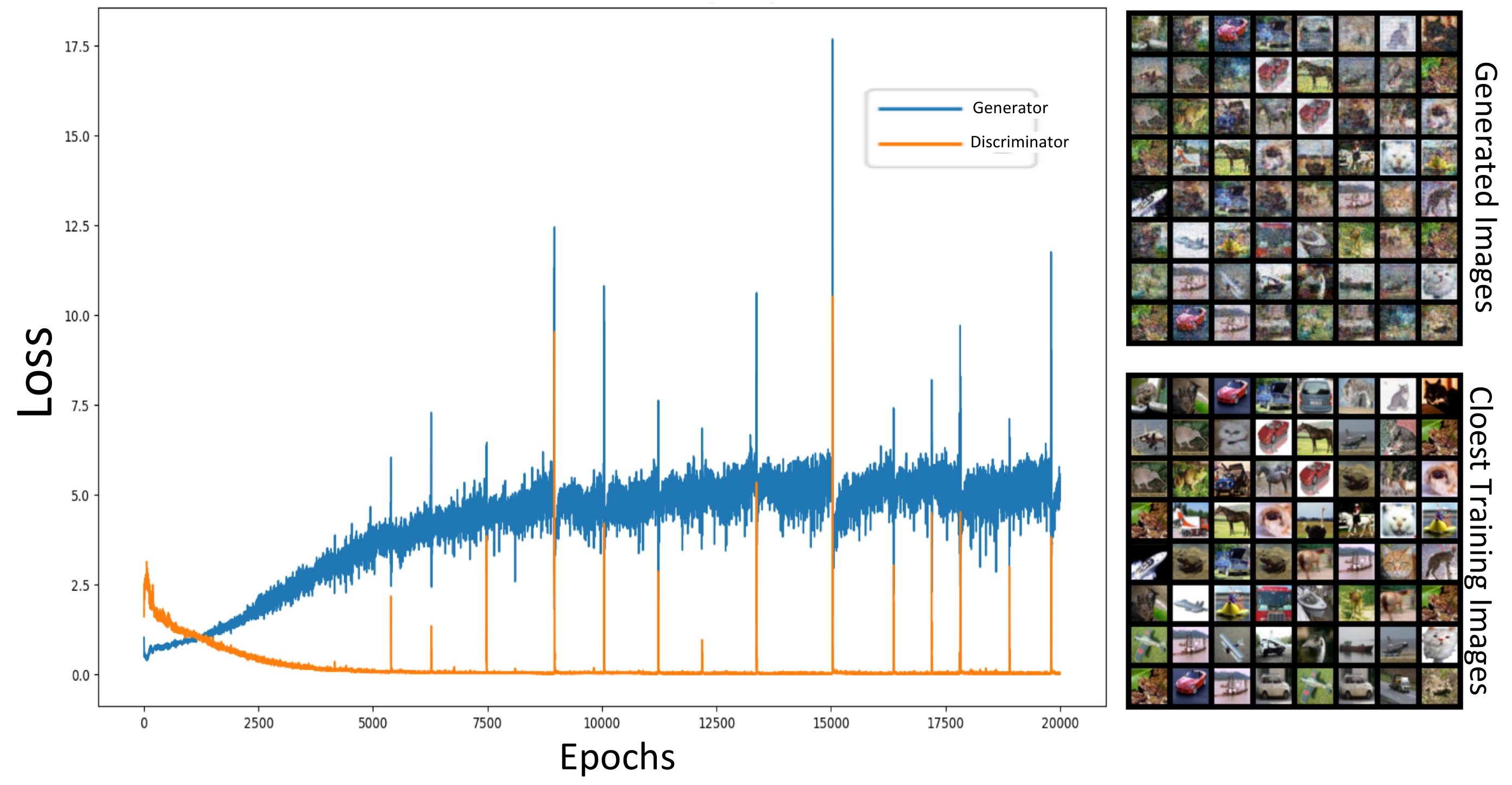}
	\DeclareGraphicsExtensions.
	\vspace{-0.6cm}
	\caption{By extending the input of generator from pure noise to optimizable mixture of Gaussians noises,
	the GANs also tend to degrade into a memorization model.}
	\label{fig_dis_gan}		
	%	\vspace{-0.4cm}
\end{figure}

The overlap coefficient in Generative Adversarial Network is 1, since the input of generator is pure noise.
In this condition,
the GAN can be considered a mapping from latent variables shared the same expectation which is usually 0.
Our idea to reduce overlap coefficient in GANs is to extend the input noise from a single distribution to a mixture distribution, such as a Gaussian mixture.
Moreover, the means of Gaussian components are also parameterized to optimizable.
During training, both the parameters of Gaussians, generator and discriminator are updated.
We utilized 1k images from CIFAR-10 for experiments, with 20000 epochs used for training.
In particular, the standard implementation in PyTorch of GAN is utilized.
Expected extending the input from pure noise into optimizable Gaussians,
all the remaining parts are kept invariant.
The resulting figure is shown in \reffig{fig_dis_gan}.
The loss of generator and discriminator are quite common compared to regular GANs,
but every generated images are very similar to training images.

\subsubsection{Memorization in Diffusion Models}
\label{sec_dis_diffusion_mem}
\begin{figure*}[ht!]
	\centering
	\includegraphics[width=\linewidth]{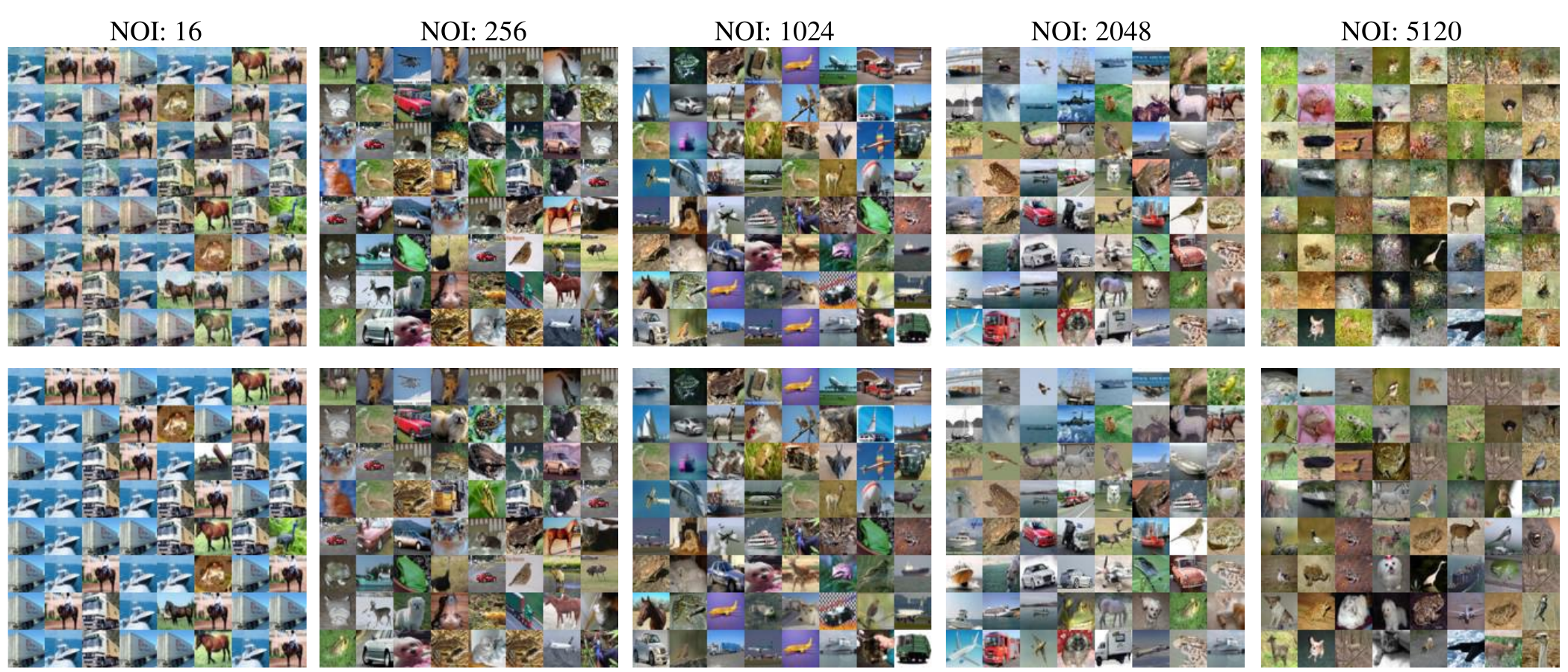}
	\caption{The DDPM model trained by different Number of Images (NOI). The images in the first row are generated images, while the images in the second row are the closest original images determined by Structural Similarity Index. We can observe that as the size of the training dataset increases, the generated images become less and less similar to the original images.}
	\label{combine_cifar}		
\end{figure*}

The easiest way to control the level of overlap coefficient is by reducing the number of training images. 
Therefore, we trained the DDPM \cite{NEURIPS2020_4c5bcfec} model on the CIFAR-10 dataset with varying numbers of training images: 16, 256, 1024, 2048, and 5120. 
To limit the fitting ability, we adopt a U-Net with only 9.27M learnable parameters as the backbone network for DDPM. 
The experiments are shown in \reffig{combine_cifar}. 
Specifically, the first row displays the generated results from trained models with varying numbers of training images: 16, 256, 1024, 2048, and 5120. 
The second row shows the corresponding similar images in the training set. 
We employ the Structural Similarity Index (SSIM) to find the most similar images. 
As shown in \reffig{combine_cifar}, when the number of training images is 16, the diffusion model always outputs training images. 
When the number of training images increases to 1024, we can only see a few differences in the details. When the training images further increase to 5120, the diffusion model demonstrates images that are close to image fusion.

\subsubsection{Overfitting Discussion}
\label{sec_dis_overfit}
The original purpose of overfitting is to describe the gap between training and testing performance, which reflects generalization ability.
In generation tasks, however, there is no single gold-standard target for a test set, although generalization can still be assessed on held-out data via proxy metrics (e.g., likelihood, FID, human evaluation).
Consequently, equating overfitting with mere memorization is intuitively appealing but not strictly correct.
Moreover, considering that the target of an autoregressive model is to fit the empirical distribution, a vanishing training loss on finite data often increases the risk of memorization.
Thus, some methods deliberately avoid over-optimization of the training objective as a form of regularization aimed at improving generalization performance.
Unfortunately, from a mathematical-logical perspective, given that a loss function is designed to measure the discrepancy between a model and its target, the natural interpretation is that the global optimum corresponds to the loss approaching zero.
In such a case, achieving zero loss should indicate that the model has perfectly captured the target distribution.
However, in practice, the notion of overfitting is often introduced to suggest that a vanishing training loss reflects memorization rather than generalization.
This raises a conceptual tension: if zero is not regarded as the true optimum, then how should one define the boundary between acceptable convergence and overfitting?
Is approaching zero asymptotically still problematic, or only reaching it exactly?
From this perspective, overfitting appears less as a logically necessary concept.

\subsubsection{FID Comparison Fairness}
\label{sec_dis_comp}
A common concern could be the fairness of comparing the proposed ARVM with state-of-the-art generative models.
Practically, it is true that such a comparison is “unfair,” since the FID results of the proposed approach essentially come from memorization, while there is clear evidence that SOTA methods like diffusion are capable of generating novel images.
However, from a mathematical-logical perspective, since no prior images are explicitly involved in the sampling steps, the FID of the proposed ARVM is still comparable.
Logically speaking, even a purely memorization-based model still fits the minimal definition of a generative model.
Of course, our goal is not to argue over semantics.
The true issue lies in the evaluation metric: FID is insensitive to memorization.
Moreover, since the proposed approach is basically an autoregressive model, there is no evidence to disprove that the FID reported by SOTA methods is not also benefiting from memorization, especially in autoregressive settings.
Indeed, numerous recent works point to memorization in various generative paradigms, including diffusion \citep{carlini2023extracting} and autoregressive models \citep{kowalczuk2025privacy,kasliwal2025localizing,yu2025icas}.
Ultimately, the main focus of the proposed approach is not to ``beat'' SOTA methods, but to encourage critical reflection on the core assumptions underlying generative modeling.

\subsection{Mathematical Derivation}
\subsubsection{Proof of Reconstruction MSE Lower Bound in \refthm{thm_lower_bound}}
\label{sec_math_proof_lower_bound}
Given a set of images $\mathbf{X}=\{\mathbf{x}_i\}_{i=1}^N$, encoder $e_{\theta}$ and decoder $d_{\phi}$.
Let $\mathbf{z}_i=e_{\theta}(\mathbf{x}_i)$ and inject symmetric, unimodal noise to obtain $\tilde{\mathbf{z}}_i=\mathbf{z}_i+\epsilon_i$ with density $p_{\tilde{\mathbf{z}}_i}(\cdot)$.
For any pair $(i,j)$ define:
\begin{equation}
p_m^{(i,j)}(\mathbf{z}) \;=\; \min\!\big(p_{\tilde{\mathbf{z}}_i}(\mathbf{z}),\,p_{\tilde{\mathbf{z}}_j}(\mathbf{z})\big).
\end{equation}
Then:
\begin{equation}
\begin{aligned}
 \frac{1}{N}\sum_{i=1}^N\mathbb{E}\big[\|d(\tilde{\mathbf{z}}_i) - \mathbf{x}_i\|^2\big] & = \frac{1}{2N^2}\sum_{i=1}^{N}\sum_{j=1}^{N} 
\Big( \mathbb{E}\|d(\tilde{\mathbf{z}}_i) - \mathbf{x}_i\|^2 + \mathbb{E}\|d(\tilde{\mathbf{z}}_j) - \mathbf{x}_j\|^2 \Big) \\[2pt]
&= \frac{1}{2N^2}\sum_{i,j} \Big( \int p_{\tilde{\mathbf{z}}_i}(\mathbf{z})\|d(\mathbf{z})-\mathbf{x}_i\|^2\,d\mathbf{z}
+ \int p_{\tilde{\mathbf{z}}_j}(\mathbf{z})\|d(\mathbf{z})-\mathbf{x}_j\|^2\,d\mathbf{z} \Big) \\[2pt]
&= \frac{1}{2N^2}\sum_{i,j} \int p_m^{(i,j)}(\mathbf{z}) \Big( \|d(\mathbf{z})-\mathbf{x}_i\|^2 + \|d(\mathbf{z})-\mathbf{x}_j\|^2 \Big)\,d\mathbf{z} \\
& \quad + \frac{1}{2N^2}\sum_{i,j} \zeta_{ij}(\mathbf{Z},\mathbf{X}),
\end{aligned}
\end{equation}
where the expression of $\zeta_{ij}(\mathbf{Z},\mathbf{X})$ is:
\begin{equation}
\zeta_{ij}(\mathbf{Z},\mathbf{X})
:= \int \big(p_{\tilde{\mathbf{z}}_i}(\mathbf{z})-p_m^{(i,j)}(\mathbf{z})\big)\|d(\mathbf{z})-\mathbf{x}_i\|^2\,d\mathbf{z}
+ \int \big(p_{\tilde{\mathbf{z}}_j}(\mathbf{z})-p_m^{(i,j)}(\mathbf{z})\big)\|d(\mathbf{z})-\mathbf{x}_j\|^2\,d\mathbf{z}.
\end{equation}
Let $A_{ij}=\{\mathbf{z}:\,p_{\tilde{\mathbf{z}}_i}(\mathbf{z})\ge p_{\tilde{\mathbf{z}}_j}(\mathbf{z})\}$ and $B_{ij}=A_{ij}^{\complement}$. 
On $A_{ij}$, $p_{\tilde{\mathbf{z}}_i}-p_m^{(i,j)}=p_{\tilde{\mathbf{z}}_i}-p_{\tilde{\mathbf{z}}_j}\ge 0$ and $p_{\tilde{\mathbf{z}}_j}-p_m^{(i,j)}=0$; on $B_{ij}$, the roles swap. 
Since the weights are nonnegative and the squared terms are nonnegative, we have:
\[
\boxed{\;\zeta_{ij}(\mathbf{Z},\mathbf{X}) \ge 0\;}.
\]
By the parallelogram inequality,
\begin{equation}
\|d(\mathbf{z})-\mathbf{x}_i\|^2+\|d(\mathbf{z})-\mathbf{x}_j\|^2 \;\ge\; \tfrac12\|\mathbf{x}_i-\mathbf{x}_j\|^2,
	\label{eqn_inequ_squre}
\end{equation}
hence:
\begin{equation}
\frac{1}{N}\sum_{i=1}^N\mathbb{E}\big[\|d(\tilde{\mathbf{z}}_i) - \mathbf{x}_i\|^2\big]
\;\ge\;
\frac{1}{4N^2}\sum_{i,j} \|\mathbf{x}_i-\mathbf{x}_j\|^2 \int p_m^{(i,j)}(\mathbf{z})\,d\mathbf{z}
\;+\; \frac{1}{2N^2}\sum_{i,j} \zeta_{ij}(\mathbf{Z},\mathbf{X}).
\end{equation}
Since $\int p_m^{(i,j)}(\mathbf{z})\,d\mathbf{z}=\mathrm{OC}(\tilde{\mathbf{z}}_i,\tilde{\mathbf{z}}_j)$ and $\zeta_{ij}\ge 0$, we obtain the lower bound
\[
\boxed{\;
\frac{1}{N}\sum_{i=1}^N\mathbb{E}\big[\|d(\tilde{\mathbf{z}}_i) - \mathbf{x}_i\|^2\big]
\;\ge\; 
\frac{1}{4N^2}\sum_{i,j=1}^{N} \mathrm{OC}(\tilde{\mathbf{z}}_i,\tilde{\mathbf{z}}_j)\,\|\mathbf{x}_i-\mathbf{x}_j\|^2
\; }.
\]

\begin{remark}
Although the above lower bound is derived under the squared error loss,
the key structure does not rely on the specific quadratic form.
The overlap coefficient $\mathrm{OC}(\tilde{\mathbf{z}}_i,\tilde{\mathbf{z}}_j)$
arises solely from the probabilistic overlap of the perturbed latent codes
and is independent of the loss.
	The constant factor $\tfrac{1}{2}||\mathbf{x}_i - \mathbf{x}_j||^2$ in the bound originates from the parallelogram inequality in \refeqn{eqn_inequ_squre},
which is a consequence of the strong convexity of the squared norm.
For a general convex (or strongly convex) loss,
one may obtain an analogous lower bound where the constant changes according to the convexity parameter of the chosen loss.
Thus, the phenomenon that reconstruction error is fundamentally limited by the overlap coefficient in MEPS is not specific to the squared loss, but extends to a broader family of convex losses.
\end{remark}

\subsubsection{Proof of Mutual Exclusivity in \refthm{thm_mutual}}
\label{sec_math_proof_mutual}
The expression of \refthm{thm_mutual} is shown as:
\begin{equation}
      \argmin_{ {\mathbf{z}}_i, {\mathbf{z}}_j}\; 
   \mathbb{E}_{\epsilon} [\mathrm{OC}(\tilde{\mathbf{z}}_i,\tilde{\mathbf{z}}_j)]
   \Rightarrow  
   \argmin_{ {\mathbf{z}}_i, {\mathbf{z}}_j}\; 
   \mathbb{E}_{\epsilon} [\mathrm{OC}({\mathbf{z}}_i + \epsilon,{\mathbf{z}}_j + \epsilon)]
   \Rightarrow  
   \argmax_{ \mathbf{z}_i, \mathbf{z}_j }\; \frac{1}{N^2} \sum_{i,j}\|\mathbf{z}_i - \mathbf{z}_j\|^2 .
\end{equation}
where noise $\epsilon$ is a symmetric, unimodal function with $f(\cdot)$ as its probability density function.
Therefore,
the probability density function of $\tilde{\mathbf{z}}_i$ and $\tilde{\mathbf{z}}_j$ is:
\begin{equation}
p_{\tilde{\mathbf{z}}_i}(\mathbf{z}) = f(\mathbf{z} - \mathbf{z}_i), \quad p_{\tilde{\mathbf{z}}_j}(\mathbf{z}) = f(\mathbf{z} - \mathbf{z}_j). 
\end{equation}
Then, by plugging this expression into the overlap coefficient in \refeqn{eqn_OC}, we have:
\begin{equation}
\begin{aligned}
   \mathrm{OC}(\tilde{\mathbf{z}}_i,\tilde{\mathbf{z}}_j) = \int \min\!\big(p_{\tilde{\mathbf{z}}_i}(\mathbf{z}),\, p_{\tilde{\mathbf{z}}_j}(\mathbf{z})\big)\, d\mathbf{z} = \int \min \left( f(\mathbf{z} - \mathbf{z}_i),\; f(\mathbf{z} - \mathbf{z}_j) \right) d\mathbf{z}.
\end{aligned}
\end{equation}
Since $f$ is symmetric and radially unimodal (e.g., Gaussian), the overlap coefficient $\mathrm{OC}(\tilde{\mathbf{z}}_i, \tilde{\mathbf{z}}_j)$ depends solely on the Euclidean distance $d_{ij} = \|\mathbf{z}_i - \mathbf{z}_j\|$. Then we have:
\begin{equation}
\mathrm{OC}(\tilde{\mathbf{z}}_i, \tilde{\mathbf{z}}_j) = h(\|\mathbf{z}_i - \mathbf{z}_j\|),
\end{equation}
where $h(\cdot)$ is a strictly decreasing function.
Therefore, minimizing the sum of all pairwise overlaps is equivalent to minimizing the sum over all $h(d_{ij})$. Since $h(\cdot)$ is strictly decreasing, this objective is effectively enforced by maximizing the pairwise distances $d_{ij} = \|\mathbf{z}_i - \mathbf{z}_j\|$. \refthm{thm_mutual} is thus proved.
\begin{remark}
This proof shows that minimizing overlap between symmetric, unimodal latent distributions is mathematically equivalent to maximizing their pairwise distances. 
The main limitation of this proof is the reliance on symmetric, unimodal assumptions, which may not extend to more complex or multimodal priors.
However, as the noise is usually injected into the elements of tensors,
such a proof is sufficient for analyzing our MEPS in variable generative models like VAEs, GANs, and diffusion.
\end{remark}

\subsubsection{On the Monotonicity of OC with Respect to Scale}

For the two distributions we use (Gaussian and symmetric triangular), 
the overlap coefficient (OC) between two shifted copies with fixed mean separation 
$\Delta$ increases monotonically with the scale parameter $\sigma$.

\subsubsubsection{\textbf{Gaussian case:}}
For $X\sim \mathcal{N}(\mu_1,\sigma^2)$ and $Y\sim \mathcal{N}(\mu_2,\sigma^2)$ 
with $\Delta=|\mu_1-\mu_2|$, the overlap coefficient is
\begin{equation}
\text{OC}(\sigma,\Delta)=2\,\Phi\!\left(-\tfrac{\Delta}{2\sigma}\right),
\end{equation}
where $\Phi$ is the standard Gaussian CDF. Differentiating gives
\begin{equation}
\frac{\partial \text{OC}}{\partial \sigma} 
= 2\,\phi\!\left(\tfrac{\Delta}{2\sigma}\right)\cdot \tfrac{\Delta}{2\sigma^2} \; > \; 0,
\end{equation}
so $\text{OC}$ increases strictly with $\sigma$.

\subsubsubsection{\textbf{Triangular case:}}
\label{sec_dis_gan_oc}
For triangular distribution, we set $\kappa=2$, centered at $\mu_1,\mu_2$ with half-width $\sigma$, 
the overlap region is the intersection of two isosceles triangles. Its area is a quadratic 
function of the overlap length, which grows linearly with $\sigma$. 
A direct calculation gives
\begin{equation}
\text{OC}(\sigma,\Delta) \;=\; 
\begin{cases}
\left(1-\tfrac{\Delta}{2\sigma}\right)^2, & 0 \le \Delta \le 2\sigma, \\[6pt]
0, & \Delta \ge 2\sigma.
\end{cases}
\end{equation}
Clearly, $\tfrac{\partial \text{OC}}{\partial\sigma} > 0$ whenever overlap exists.
\begin{remark}
Thus, in both Gaussian and triangular settings, scaling $\sigma$ monotonically 
enlarges the overlap for fixed $\Delta$, justifying our use of $\sigma$ 
as a practical proxy for controlling OC.
\end{remark}

\subsubsection{OC=1 in GANs}
For GANs, the generator input is pure noise. During training, the number of input noise vectors usually equals the batch size, and each noise vector is mapped through the generator to produce a fake image.
In our MEPS framework, these input noise vectors can be regarded as a set of random variables. Specifically, drawing n
samples from the same distribution is equivalent to defining n random variables that follow the same distribution with identical expectation and sampling each once. Under this view, all input variables in GANs share the same distribution and expectation, and thus their overlap coefficient (OC) equals 1.
This situation corresponds to an extreme case in MEPS where all random variables completely overlap. Intuitively, this means that the model lacks any separation margin during training, making the optimization more unstable. We believe this perspective offers an explanation for the well-known training difficulties of GANs. It should be emphasized that this is not a formal proof, but rather an interpretative understanding.

\subsection{Diagnostic Evaluation: Comparison with State-of-the-Art Methods}
\label{sec_cmp_ARVM}
We also compare our $\gamma$-Autoregressive Random Variable Model (ARVM), with observation range of 7, 5, and 3,
to get the FID scores of $7$-ARVM, $5$-ARVM and $3$-ARVM shown in \reftab{tab_comp_cifar} and \reftab{tab_comp_lsun_imagenet_celebahq}.
In particular, the architecture is described in \reftab{tab_net_bl_arvm}, with NoD=32.
Since the spatial size of binary latent is $4 \times 4$, our ARVM learns the global distribution when observation range = 7 (padding is used when spatial size is too small.)
$7$-ARVM achieves an FID score of 0.56 when learning global distributions, which often results in training-sample memorization.
Unfortunately, such FID scores remain comparable to those of state-of-the-art methods under identical evaluation conditions (\refsec{sec_dis_comp}).
Notably, this is achieved without relying on any prior assumptions related to image structure during the sampling process.
Similar results are observed on high-resolution datasets including ImageNet, CelebA-HQ, and LSUN Bedroom, as shown in \reftab{tab_comp_lsun_imagenet_celebahq}.
The main reason for such low FID is primarily the memorization effect in the proposed $\gamma$-ARVM.
However, since the proposed ARVM is essentially a standard autoregressive model, especially when the observation range is increased to learn global distributions,
it is worth considering that the current claims that autoregressive models outperform diffusion models may simply be a consequence of memorization \citep{sun2024autoregressive,zhang2025diffusion}.
Likewise, it is also worth considering that the reported superiority of diffusion over VAEs or GANs may be due to the same reason.

\begin{table}[ht!]
\begin{center}
%\begin{wraptable}{r}{0.56\linewidth}
\caption{Diagnostic Evaluation on CIFAR-10 Using FID and Inception Scores on the CIFAR-10 dataset.}
\label{tab_comp_cifar}
%\scalebox{1.0}{
%\resizebox{1.0}{
\begin{tabular}{l|l|rr}
\toprule
& \makecell{\textbf{Method}} & \makecell{\textbf{FID score} $\downarrow$} & \makecell{\textbf{Inception Score}$\uparrow$} \\
\midrule\midrule
\multirow{2}{*}{\textbf{Diffusion}} & DDPM \citep{ho2020denoising} & 3.17 & 9.46 $\pm$ 0.11 \\
& EDM \citep{cui2023kd} & 1.30 & N/A \\
\midrule
\multirow{5}{*}{\textbf{GAN}} & CCF-GAN \citep{li2023neural} & 6.08 & N/A \\
& KD-DLGAN \citep{cui2023kd} & 8.30 & N/A \\
%& BigGAN \citep{brock2018large} & 14.73 & 9.22 \\
& StyleGAN2 \citep{karras2020training} & 3.26 & 9.74 $\pm$ 0.05 \\
& SN-SMMDGAN \citep{arbel2018gradient} & 25.00 & 7.30 \\
\midrule
\multirow{5}{*}{\textbf{VAE}} & NCP-VAE \citep{aneja2021contrastive} & 24.08 & N/A \\
& NVAE \citep{vahdat2020nvae} & 32.53 & N/A \\
& DC-VAE \citep{parmar2021dual} & 17.90 & 8.20 \\
%& QSNGAN \citep{grassucci2022quaternion} & 31.96 & 4.71 \\
& NCSN  \citep{song2019generative} & 25.32 & 8.87 \\
\midrule
\multirow{3}{*}{\textbf{Ours}} & ARVM$_3$ & 67.13 & 6.32 $\pm$ 0.23 \\
& ARVM$_2$ & 31.42 & 7.12 $\pm$ 0.15 \\
	& ARVM$_1$ & 0.56 &  11.15 $\pm$ 0.13 \\
\bottomrule
\end{tabular}
%}
\end{center}
\end{table}

\begin{table}[ht!]
%\begin{wraptable}{r}{0.56\linewidth}
	\caption{Diagnostic Evaluation on CIFAR-10 Using FID and Inception Scores on high-resolution datasets.}
	\label{tab_comp_lsun_imagenet_celebahq}
	\centering
%	\scalebox{0.75}{
		\begin{tabular}{l|l|l|r}
			\toprule
			\textbf{Dataset} & \makecell{\textbf{Model}} & \makecell{\textbf{Method}} & \makecell{\textbf{FID} $\downarrow$} \\
			\midrule\midrule
			\multirow{5}{*}{\textbf{\makecell{LSUN \\ Bedroom}}}
			& \textbf{Diffusion} & DDPM \citep{ho2020denoising} & 6.36 \\
			\cmidrule{2-4}
			& \multirow{2}{*}{\textbf{GAN}} & PGGAN \citep{karras2018progressive} & 8.34 \\
			& & PG-SWGAN \citep{wu2019sliced} & 8.00 \\
			\cmidrule{2-4}
			& \multirow{1}{*}{\textbf{Ours}} & ARVM$_{1}$ & 1.54 \\
			\midrule
			\multirow{6}{*}{\textbf{ImageNet}} 
			& \multirow{2}{*}{\textbf{Diffusion}} & DiT-XL/2 \citep{peebles2023scalable} & 9.62 \\
			& & DiT-XL/2-G \citep{peebles2023scalable} & 2.27 \\
			\cmidrule{2-4}
			& \multirow{2}{*}{\textbf{Transformer}} & MaskGIT \citep{chang2022maskgit} & 6.18 \\
			& & VQGAN+Transformer \citep{esser2021taming} & 6.59 \\
			\cmidrule{2-4}
			& \multirow{1}{*}{\textbf{Ours}} & ARVM$_{1}$ & 5.63 \\
			\midrule
			\multirow{3}{*}{\textbf{\makecell{CelebA-HQ \\ 256x256}}}
			& \multirow{1}{*}{\textbf{VAE}} & NVAE \citep{vahdat2020nvae} & 48.27 \\
			\cmidrule{2-4}
			& \multirow{1}{*}{\textbf{Ours}} & ARVM$_{1}$ & 1.53 \\
			\bottomrule
		\end{tabular}
%	}
%	\vspace{-10pt}
\end{table}

\subsubsection{Experimental Details}
All experiments were conducted on a single RTX 4090 GPU with 24 GB of VRAM. Training and testing for each experiment were completed within 24 hours on this single GPU, given the computational constraints. For the same reason, large-scale experiments on larger models were not feasible. All implementations were based on PyTorch, and the Adam optimizer was used for training. Source code and running scripts will be released upon acceptance of this paper.

\begin{table}[!t]
\caption{Details of our network architecture}
\label{tab_net_architect_ME}
\scriptsize
\centering
\begin{tabular}{|l|l|l|l|l|l|}
\toprule
 & Type & weight & stride & padding & Data size \\
\midrule\midrule
\parbox[t]{2mm}{\multirow{8}{*}{\rotatebox[origin=c]{90}{Encoder}}}
& Input  &  &  &  &  $N \times 3 \times 32 \times 32$  \\
& Conv2d
& $64  \times  3  \times  4  \times  4$
& \text{2}
& \text{1}
& $N  \times  64  \times  16  \times  16$ \\
& LeakyReLU
&
&
&
& \\
& Conv2d
& $256  \times  64  \times  4  \times  4$
& \text{2}
& \text{1}
& $N  \times  256  \times  8  \times  8$ \\
& LeakyReLU
&
&
&
& \\
& Conv2d
& $256  \times  1024 \times  1 \times  1 $
& 1
& 0
& $N  \times  1024  \times  8  \times  8$ \\
& Conv2d
& $1024  \times  \text{NoD}\times  1 \times  1  $
& 1
& 0
& $N  \times  \text{NoD}  \times  8  \times  8$ \\
\midrule
\parbox[t]{2mm}{\multirow{3}{*}{\rotatebox[origin=c]{90}{Latents}}}
&   &  &  &  &   \\
&   &  &  &  & $N \times \text{NoD} \times 8 \times 8$  \\
&   &  &  &  &   \\
\midrule
\parbox[t]{2mm}{\multirow{11}{*}{\rotatebox[origin=c]{90}{Decoder}}}
& Linear
& $\text{NoD}  \times  1024\times  1 \times  1 $
&\text{1}
& \text{0}
& $N  \times  1024  \times  8  \times 8$ \\
& Linear
& $1024  \times  1024 \times  1 \times  1 $
&\text{1}
& \text{0}
& $N  \times  1024  \times  8  \times  8$ \\
& LeakyReLU
&
&
&
& \\
& ConvT2d
& $512  \times  1024  \times  3  \times  3$
& \text{3}
& \text{1}
& $N  \times  512  \times  8  \times  8$ \\
& LeakyReLU
&
&
&
& \\
&  ConvT2d
& $64  \times  512  \times  4  \times  4$
& \text{2}
& \text{1}
& $N  \times  64  \times  16  \times  16$ \\
&  ConvT2d
& $3  \times  64  \times  4  \times  4$
& \text{2}
& \text{1}
& $N  \times  3  \times  32  \times  32$ \\
& Tanh
&
&
&
&\\
\midrule
\parbox[t]{2mm}{\multirow{4}{*}{\rotatebox[origin=c]{90}{Refine}}}
& Conv2d
& $32  \times  3  \times  1  \times  1$
&\text{3}
& \text{1}
& $N  \times  32  \times  32  \times  32$ \\
& LeakyReLU
& $\alpha = 0.01$
&
&
& $N  \times  32  \times  32  \times  32$\\
& Conv2d
& $3  \times  32  \times  1  \times  1$
&\text{3}
& \text{1}
& $N  \times  3  \times  32  \times  32$ \\
\midrule
& Output
&
&
&
& $N  \times  3  \times  32  \times  32$ \\
\bottomrule
\end{tabular}
\\
\scriptsize NoD: number of dimension. \ \ \ \ \ \ \ \ \ \ \ \ \ \ \ \ \ \ \ \ \ \ \ \ \ \ \ \ \ \ \ \ \ \ \ \ \ \ \ \ \ \ \ \ \ \ \ \ \ \ \ \ \ \ \ \ \ \ \ \ \ \ \ \ \ \ \ \ \ \ \ \ \ \ \ \ \ \ \ \ \ \ \ \ \ \ \ \ \ \ \ \ \ \ \ \ \ \ \ \ \ \ \ \ \ \ \ \ \ \ \ \ \ \ \ 
\end{table}

\begin{table}[!t]
\caption{Details of our network architecture.}
\label{tab_net_bl_arvm}
\scriptsize
\centering
\begin{tabular}{|l|l|l|l|l|l|}
\toprule
 & Type & weight & stride & padding & Data size \\
\midrule\midrule
\parbox[t]{2mm}{\multirow{8}{*}{\rotatebox[origin=c]{90}{Encoder}}}
& Input  &  &  &  &  $N \times 3 \times 32 \times 32$  \\
& Conv2d
& $64  \times  3  \times  4  \times  4$
& \text{2}
& \text{1}
& $N  \times  64  \times  16  \times  16$ \\
& LeakyReLU
&
&
&
& \\
& Conv2d
& $256  \times  64  \times  4  \times  4$
& \text{2}
& \text{1}
& $N  \times  256  \times  8  \times  8$ \\
& LeakyReLU
&
&
&
& \\
& Conv2d
& $512  \times  256  \times  4  \times  4$
& \text{2}
& \text{1}
& $N  \times  512  \times  4  \times  4$ \\
& LakyReLU
&
&
&
&\\
& Conv2d
& $512  \times  8196 \times  1 \times  1 $
& 1
& 0
& $N  \times  8196  \times  4  \times  4$ \\
& Conv2d
& $8196  \times  \text{NoD}\times  1 \times  1  $
& 1
& 0
& $N  \times  \text{NoD}  \times  4  \times  4$ \\
\midrule
\parbox[t]{2mm}{\multirow{3}{*}{\rotatebox[origin=c]{90}{Latents}}}
&   &  &  &  &   \\
&   &  &  &  & $N \times \text{NoD} \times 4 \times 4$  \\
&   &  &  &  &   \\
\midrule
\parbox[t]{2mm}{\multirow{11}{*}{\rotatebox[origin=c]{90}{Decoder}}}
& Linear
& $\text{NoD}  \times  8196\times  1 \times  1 $
&\text{1}
& \text{0}
& $N  \times  8196  \times  4  \times 4$ \\
& Linear
& $8196  \times  1024 \times  1 \times  1 $
&\text{1}
& \text{0}
& $N  \times  1024  \times  4  \times  4$ \\
& LeakyReLU
&
&
&
& \\
& ConvT2d
& $512  \times  1024  \times  4  \times  4$
& \text{1}
& \text{0}
& $N  \times  512  \times  4  \times  4$ \\
& LeakyReLU
&
&
&
& \\
&  ConvT2d
& $256  \times  512  \times  4  \times  4$
& \text{2}
& \text{1}
& $N  \times  256  \times  8  \times  8$ \\
&  ConvT2d
& $64  \times  256  \times  4  \times  4$
& \text{2}
& \text{1}
& $N  \times  64  \times  16  \times  16$ \\
&  ConvT2d
& $3  \times  64  \times  4  \times  4$
& \text{2}
& \text{1}
& $N  \times  3  \times  32  \times  32$ \\
& Tanh
&
&
&
&\\
\midrule
\parbox[t]{2mm}{\multirow{4}{*}{\rotatebox[origin=c]{90}{Refine}}}
& Conv2d
& $32  \times  3  \times  1  \times  1$
&\text{3}
& \text{1}
& $N  \times  32  \times  32  \times  32$ \\
& LeakyReLU
& $\alpha = 0.01$
&
&
& $N  \times  32  \times  32  \times  32$\\
& Conv2d
& $3  \times  32  \times  1  \times  1$
&\text{3}
& \text{1}
& $N  \times  3  \times  32  \times  32$ \\
\midrule
& Output
&
&
&
& $N  \times  3  \times  32  \times  32$ \\
\bottomrule
\end{tabular}
\\
\scriptsize NoD: number of dimension. \ \ \ \ \ \ \ \ \ \ \ \ \ \ \ \ \ \ \ \ \ \ \ \ \ \ \ \ \ \ \ \ \ \ \ \ \ \ \ \ \ \ \ \ \ \ \ \ \ \ \ \ \ \ \ \ \ \ \ \ \ \ \ \ \ \ \ \ \ \ \ \ \ \ \ \ \ \ \ \ \ \ \ \ \ \ \ \ \ \ \ \ \ \ \ \ \ \ \ \ \ \ \ \ \ \ \ \ \ \ \ \ \ \ \ \
\end{table}

\begin{table}[ht!]
	\caption{Min-Max normalization parameters settting.}
	\label{table_details_norm}
\begin{tabularx}{\linewidth}{l|r|r|r|r}
\toprule
\textbf{ }  & \textbf{CIFAR-10 Tri.} & \textbf{MNIST Tri.} & \textbf{CIFAR-10 Gau.}  & \textbf{MNIST Gau.}\\
\midrule
min meanDist & 28.01 & 36.11& 50.76 & 72.43 \\
max meanDist & 1018.79 & 1131.68& 1302.75 & 1068.97 \\
\midrule
min minDist & 25.78 &25.26 & 23.72 & 20.11 \\
max minDist & 806.93 & 315.31& 988.17 & 414.71 \\
\midrule
min recMSE & 6.97 & 2.18& 6.43  & 1.89 \\
max recMSE & 296.37 &36.31 & 395.81 & 45.13 \\
\bottomrule
\end{tabularx}
\\
\scriptsize Tri. Triangular noise, Gau. Gaussian noise. \ \ \ \ \ \ \ \ \ \ \ \ \ \ \ \ \ \ \ \ \ \ \ \ \ \ \ \ \ \ \ \ \ \ \ \ \ \ \ \ \ \ \ \ \ \ \ \ \ \ \ \ \ \ \ \ \ \ \ \ \ \ \ \ \ \ \ \ \ \ \ \ \ \ \ \ \ \ \ \ \ \ \ \ \ \ \ \ \ \ \ \ \ \ \ \ \ \ \ \ \ \ \ \ \ \ \ \ \ \ \ \ \ \ \ \
\end{table}

\subsection{Disclosure of Large Language Models}
\label{sec_use_LLM}
We utilized Grammarly and ChatGPT solely to check typos and grammar in the proposed paper. No technical content, experiments, or analysis were generated by large language models.

\end{document}